\documentclass[
twocolumn,
hf,
]{ceurart}

\usepackage{epsfig}
\usepackage{graphicx}
\usepackage{amsmath}
\usepackage{amssymb}

\usepackage{comment}
\usepackage{color}
\usepackage{gensymb}
\usepackage[caption=false]{subfig}
\usepackage[american]{babel}
\usepackage{makecell}
\usepackage{overpic}
\usepackage{ragged2e}

\usepackage{listings}
\usepackage{booktabs}
\usepackage{multirow}
\usepackage{bbm}
\usepackage{tabularx}

\newcommand{\etal}{\textit{et al}.}
\newcommand{\ie}{\textit{i}.\textit{e}.}
\newcommand{\eg}{\textit{e}.\textit{g}.}
\newcommand{\myparagraph}[1]{\vspace{2pt}\noindent{\bf #1}}
\DeclareMathOperator*{\argmax}{arg\,max}
\DeclareMathOperator*{\argmin}{arg\,min}

\newcolumntype{P}[1]{>{\centering\arraybackslash}p{#1}}
\newcolumntype{M}[1]{>{\centering\arraybackslash}m{#1}}
\begin{document}

\copyrightyear{2023}
\copyrightclause{Copyright for this paper by its authors.
  Use permitted under Creative Commons License Attribution 4.0
  International (CC BY 4.0).}

\conference{26th Computer Vision Winter Workshop, Robert Sablatnig and Florian Kleber (eds.), Krems, Lower Austria, Austria, Feb. 15-17, 2023}

\title{TAEC: Unsupervised Action Segmentation with Temporal-Aware Embedding and Clustering}


\author[1,2]{Wei Lin}[%
email=wei.lin@icg.tugraz.at,
]
\address[1]{Institute of Computer Graphics and Vision, Graz University of Technology, Austria}
\address[2]{Christian Doppler Laboratory for Semantic 3D Computer Vision, Austria }

\author[3]{Anna Kukleva}[%
email=akukleva@mpi-inf.mpg.de,
]
\address[3]{Max-Planck-Institute for Informatics, Germany}

\author[1]{Horst Possegger}[%
email=possegger@icg.tugraz.at,
]

\author[4]{Hilde Kuehne}[%
email=kuehne@uni-frankfurt.de,
]
\address[4]{Goethe University Frankfurt, Germany}

\author[1]{Horst Bischof}[%
email=bischof@icg.tugraz.at,
]



\begin{abstract}
Temporal action segmentation in untrimmed videos has gained increased attention recently. However, annotating action classes and frame-wise boundaries is extremely time consuming and cost intensive, especially on large-scale datasets. To address this issue, we propose an unsupervised approach for learning action classes from untrimmed video sequences. In particular, we propose a temporal embedding network that combines relative time prediction, feature reconstruction, and sequence-to-sequence learning, to preserve the spatial layout and sequential nature of the video features. A two-step clustering pipeline on these embedded feature representations then allows us to enforce temporal consistency within, as well as across videos. Based on the identified clusters, we decode the video into coherent temporal segments that correspond to semantically meaningful action classes. Our evaluation on three challenging datasets shows the impact of each component and, furthermore, demonstrates our state-of-the-art unsupervised action segmentation results.
\end{abstract}

\begin{keywords}
  Unsupervised learning \sep
  unsupervised clustering \sep
  action segmentation
\end{keywords}

\maketitle

\section{Introduction}
Action recognition has seen tremendous success in recent years, especially in the context of short video clip classification~\cite{wang2021action, feichtenhofer2020x3d, yang2020temporal}, action detection~\cite{xu2020g, zhao2020bottom, bai2020boundary}, and temporal action segmentation~\cite{farha2019ms,  li2020ms, wang20boundary,wang2021temporal}.
The top-performing methods for temporal action segmentation, however, require frame-wise annotations, which is expensive and impractical for large-scale datasets \cite{farha2019ms,  li2020ms, wang20boundary,huang2020improving}.
Consequently, a large body of research focuses on weakly-supervised approaches where only an ordered list \cite{chang2019d3tw, li19weakly, richard2017weakly, richard2018neuralnetwork, huang16connectionist} or an unordered set \cite{fayyaz2020sct, li20set, richard2018action} of action labels is needed. 
These approaches assume that the actions that occur in each training video are known, and sometimes even require their exact ordering. 
Acquiring such ordered action lists, however, can still be time consuming or even infeasible. 
For applications like indexing large video datasets or human behavior analysis in neuroscience or medicine, it is often unclear what actions should be annotated. In these cases, it is necessary to automatically discover and identify recurring actions in large video datasets.

To address this problem, Sener and Yao \cite{sener18} proposed the task of unsupervised action segmentation to identify patterns of recurring actions in long, untrimmed video sequences that correspond to semantically meaningful action classes. 
Recent approaches for unsupervised action segmentation, \eg~\cite{sener18, kukleva19, vidalmata2020joint}, focus on three aspects: (1)~finding a suitable embedding space for the video data, (2)~identifying clusters of temporal segments across a large amount of videos, and (3)~parsing the input videos given the respective feature embedding and cluster information.
Sener and Yao \cite{sener18} tackle the action segmentation task with a linear embedding and Mallow decoding, while other approaches follow the pipeline of an MLP \cite{kukleva19} or U-Net embedding \cite{vidalmata2020joint}, K-means clustering, and Viterbi decoding.
However, these methods do not fully leverage the temporal relationships of frames within a video, either neglecting this information when learning the embedding~\cite{sener18, vidalmata2020joint} or when clustering \cite{kukleva19, vidalmata2020joint}. 

\begin{figure*}[!tb]
\centering
\includegraphics[width=0.9\textwidth]{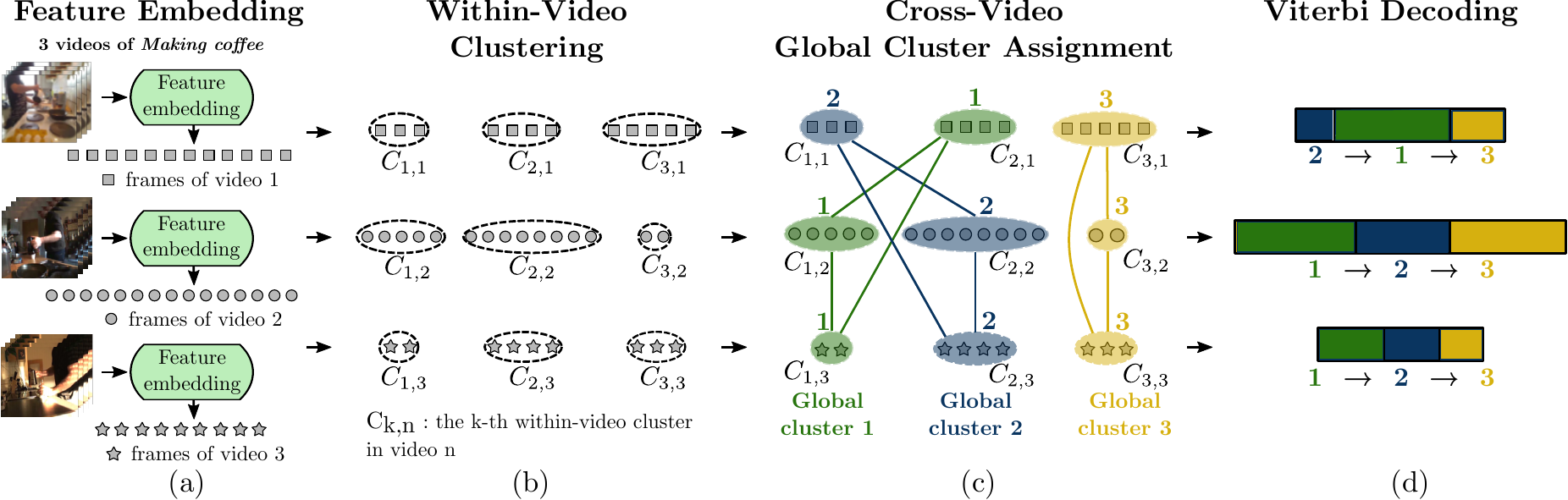}
\caption{Pipeline of TAEC. We compute the embedded features with the sequence-to-sequence temporal embedding network (SSTEN). 
Thereupon, we perform a within-video clustering on each video individually and apply a cross-video global cluster assignment to group the within-video clusters into global clusters. The global cluster assignment also defines the ordering of the clusters in each video. Finally, we use Viterbi decoding to estimate temporally coherent segments for each video.
}
\label{fig:workflow}
\end{figure*}

Since temporal consistency is essential for all steps of the action segmentation pipeline, we propose TAEC, an approach that considers the sequential nature of frames in a video for both, embedding learning and clustering. The main steps of our approach are illustrated in Fig.~\ref{fig:workflow}. 

Specifically, we first propose a sequence-to-sequence temporal embedding network (SSTEN) that combines pretext tasks of relative timestamp prediction and autoencoder feature reconstruction.
While the autoencoder reconstruction retains the feature layout, the relative timestamp prediction encodes the relative temporal information within a video.  
Sequence-to-sequence learning enables the embedding of spatial layout and temporal information on a complete video sequence. 

To cluster the embedded features, we propose a temporal-aware two-step clustering approach that consists of a within-video clustering and a cross-video global cluster assignment. 
Specifically, we perform clustering within each video, with a spatio-temporal similarity among frames. Then we conduct global cluster assignment to group the clusters across videos. The global cluster assignment defines the ordering of the clusters for each video.
In this way, we overcome the unrealistic assumption that actions of an activity always follow the same temporal order. Such an assumption is commonly used in related works, \eg~\cite{kukleva19,vidalmata2020joint}.
For instance, in the activity of \textit{making coffee}, a unified temporal order between actions such as \textit{adding milk} and \textit{adding sugar} is assumed for all videos of \textit{making coffee}, whereas our approach can handle changes of the action order in different videos. After assigning all within-video clusters to a set of global clusters, we perform Viterbi decoding to obtain a segmentation of temporally coherent segments. 


Our contributions can summarized as following:
\begin{itemize}
    \item We design a sequence-to-sequence temporal embedding network (SSTEN), which combines relative timestamp prediction, autoencoder reconstruction and sequence-to-sequence learning.
    \item We propose a within-video clustering with a novel spatio-temporal similarity formulation among frames. 
    \item We propose a cross-video global cluster assignment to group within-video clusters across videos into global clusters, which also overcomes the assumption that in all videos of an activity, actions follow the same temporal order. 
\end{itemize}



\section{Related Work}
\myparagraph{Unsupervised learning of video representations} is commonly performed via pretext tasks, such as reconstruction \cite{bhatnagar2017unsupervised, srivastava2015unsupervised}, future frame prediction \cite{vidalmata2020joint, denton2017unsupervised, villegas2017decomposing}, and recognition of frame orders \cite{kim2019self, lee2017unsupervised, Ramanathan_2015_ICCV, Fernando_2015, cherian2017generalized}. For instance, Srivastava \etal~\cite{srivastava2015unsupervised} exploit an LSTM-based autoencoder for learning video representations. Villegas \etal~\cite{villegas2017decomposing} and Denton and Birodkar \cite{denton2017unsupervised} employed two encoders to generate feature representations of content and motion. The temporal order of frames or small chunks is utilized as a self-supervision signal for representation learning on short video clips in \cite{kim2019self} and \cite{lee2017unsupervised}. Inspired by these approaches, we employ two self-supervision tasks: feature reconstruction and relative time prediction. 

\myparagraph{Clustering of temporal sequences} has been explored for parsing human motions \cite{hoai2012maximum,li15, tierney2014subspace, zhang2018human}. 
While Zhang \etal~\cite{zhang2018human} proposed a hierarchical dynamic clustering framework, Li \etal~\cite{li15} and Tierney \etal~\cite{tierney2014subspace} explored temporal subspace clustering to segment human motion data.
In contrast to unsupervised action segmentation, these methods are applied on each temporal sequence individually and do not consider association among sequences.
Instead, we propose a cross-video global cluster assignment to group within-video clusters across different videos into global clusters.

\myparagraph{Unsupervised action segmentation} on fine-grained activities has recent work that either focus on the representation learning~\cite{sener18, vidalmata2020joint, alayrac2016unsupervised} or the clustering step~\cite{bhatnagar2017unsupervised}.
However, the temporal information is neglected in at least one of these two steps. For representation learning, Sener and Yao~\cite{sener18} construct a feature embedding by learning a linear mapping from visual features to a latent space with a ranking loss. However, the linear model trained with individual frames does not consider the temporal association between frames. VidalMata \etal~\cite{vidalmata2020joint} employ a U-Net trained on individual frames for future frame prediction. Predicting for one or a few steps ahead only requires temporal relations within a small temporal window. Instead, we propose to learn a representation by predicting the complete sequence of relative timestamps to encode the long-range temporal information.

For the clustering step, related works \cite{vidalmata2020joint, bhatnagar2017unsupervised} neglect temporal consistency of frames within a video. Instead, we apply within-video clustering on each video with a proposed similarity formulation that considers both spatial and temporal distances.

Two recent approaches perform clustering \cite{sarfraz2021temporally} or cluster-agnostic boundary detection \cite{aakur19} on each video separately, without identifying clusters or segments across videos. \cite{sarfraz2021temporally} solves a task similar to human motion parsing and evaluates the segmentation for each video individually. 
\cite{aakur19} only detects boundaries of category-agnostic segments, and does not identify if some segments within a video or across videos are of the same category. On the contrary, our segments on all videos are category-aware as they are aligned globally across videos by our global cluster assignment.  

\section{Temporal-Aware Embedding and Clustering (TAEC)}
We address unsupervised action segmentation as illustrated in Fig.~\ref{fig:workflow}. 
First, we learn a suitable feature embedding (Sec.~\ref{sec:feature_embedding}). 
We then perform within-video clustering on each video (Sec.~\ref{sec:video_wise_clustering}), and group the within-video clusters into global clusters (Sec.~\ref{sec:global_cluster_assignment}). Finally, we compute temporally coherent segments on each video using Viterbi decoding (Sec.~\ref{sec:frame_labeling}).

\subsection{SSTEN: Sequence-to-Sequence Temporal Embedding Network}\label{sec:feature_embedding}
\begin{figure}[t]
\centering
\includegraphics[width=0.9\linewidth]{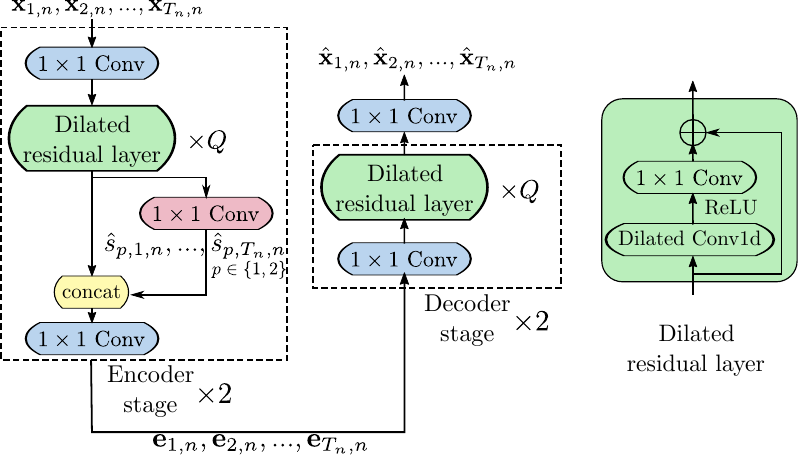}
\caption{Architecture of SSTEN: We stack 2 stages of encoder and decoder for sequence-to-sequence feature reconstruction. Each stage consists of $Q$ dilated residual layers with dilated temporal convolution. The intermediate representation in encoder is used for relative time prediction (red block).}
\label{fig:network_structure}
\end{figure}

To learn a latent representation for temporal sequences, we adopt a sequence-to-sequence autoencoder.
Inspired by the multi-stage temporal convolutional network~\cite{farha2019ms}, we use a concatenation of two stages for both encoder and decoder, as shown in Fig.~\ref{fig:network_structure}. 
Given a set $\{ \mathbf{X}_n \}_{n=1}^N$ of $N$ videos, where each video $\mathbf{X}_n=\{\mathbf{x}_{t,n}\}_{t=1}^{T_n}$ has $T_n$ frames, the outputs are reconstructed frame features $\{\hat{\mathbf{x}}_{t,n}\}_{t=1}^{T_n}$.
The embedded features are the hidden representation $\{\mathbf{e}_{t,n}\}_{t=1}^{T_n}$.

Every encoder and decoder stage consist of $1\times1$ convolution layers for dimension adjustment (Fig.~\ref{fig:network_structure} blue) and $Q$ dilated residual layers (green), each containing a dilated temporal 1D convolution. Since no fully connected layers are employed, sequences of variable lengths can be processed seamlessly. The dilation rate at the $q$-th layer is $2^{q-1}$. By stacking dilated residual layers, the temporal receptive field increases exponentially. The receptive field of the $q$-th layer is $1+(r-1)\times(2^q-1)$, where $r$ is the kernel size.
Therefore, each frame in the hidden representation has a long temporal dependency on the input video. 
In each encoder stage, we use a $1\times1$ convolution layer (in red) to predict the frame-wise relative timestamps $s_{t,n} = \frac{t}{T_n}$. At the end of each encoder stage, the hidden representation is a concatenation (in yellow) of the features from dilated residual layers and the predicted relative timestamps. 
The training loss is:
\begin{equation}\label{eq:recontr_loss_and_reltime_loss}
\scriptsize
    \mathcal{L} = \lambda \sum_{n=1}^{N} \sum_{t=1}^{T_n} \left\lVert \mathbf{x}_{t,n} - \hat{\mathbf{x}}_{t,n} \right\rVert^2_2 + \sum_{p\in\{1,2\}} \sum_{n=1}^{N} \sum_{t=1}^{T_n} ( s_{t,n} - \hat{s}_{p,t,n})^2,
\end{equation}
where the coefficient $\lambda$ balances the two terms. 
The pretext tasks of reconstruction and relative timestamp prediction encode both, the spatial distribution and the global temporal information, into the embedded features.
We compare SSTEN with several baseline embedding networks in the supplementary. 


\subsection{Two-Step Clustering}\label{sec:introducing_clustering}
After learning the feature embedding, we group the embedded features into $K$ clusters by a within-video clustering and a cross-video global cluster assignment.



\subsubsection{Within-Video Clustering}\label{sec:video_wise_clustering}
We perform spectral clustering on frames within each video (detailed description in the supplementary). 
Given the embedded feature sequence\footnote{For ease of notation, we omit the video index $n$.} $[\mathbf{e}_1, \mathbf{e}_2,...,\mathbf{e}_T]$, we build a frame-to-frame similarity matrix $G\in \mathbb{R}^{T\times T}$. The entries $g(i,j),\ i,j\in\{1,..., T\}$, represent the similarity between frame $i$ and frame $j$. 
To consider both the spatial and temporal distance of features, we propose to measure the similarity by the product of two Gaussian kernels
\begin{equation}\label{eq:spatiotemporal_similarity}
    g(i,j) = \exp{\!\left(-\frac{\left\lVert \mathbf{e}_i - \mathbf{e}_j \right\rVert^2_2}{\sigma_{\text{spat}}^2}\right)} 
    \cdot 
    \exp{\!\left(-\frac{(s_i - s_j)^2}{\sigma_{\text{tmp}}^2}\right)},
\end{equation}
where $s_i, s_j$ are the corresponding relative timestamps of frame $i,j$ and $\sigma_{\text{spat}},\sigma_{\text{tmp}}$ are the scaling factors for the spatial and temporal Gaussian kernels.
To avoid manually tuning $\sigma_{\text{spat}}$, we use local scaling \cite{zelnik2005self} to estimate $\sigma_{\text{spat}}$ dynamically. To this end, we replace $\sigma_{\text{spat}}^2$ by $\sigma_i\sigma_j$, where $\sigma_i$ is the distance from $\mathbf{e}_i$ to its \textbf{$m$}-th nearest neighbor in the embedding space. We provide an ablation study on scaling of the spatio-temporal similarity in the supplementary.
Consequently, frames of similar visual content and relative timestamps are encouraged to be grouped into the same cluster.


\subsubsection{Cross-Video Global Cluster Assignment}\label{sec:global_cluster_assignment}
After within-video clustering, we assign the $N\times K$ within-video clusters across videos into $K$ global clusters. Every global cluster should contain $N$ within-video clusters, each coming from a different video (\textit{c.f.}, Fig.~\ref{fig:workflow}). This can be interpreted as an $N$-dimensional assignment problem~\cite{pierskalla1968letter}. 

We regard the $n$-th video $V_n=\{\mathbf{c}_{k,n}|k=1,..,K\}$ as a vertex set, where each $k$-th within-video cluster $\mathbf{c}_{k,n}$ is a vertex.
We construct an $N$-partite graph $G=(V_1\cup V_2\cup ... \cup V_N, E)$. $E=\bigcup_{m<n, m,n\in\{1,...,N\}} \{(\mathbf{c},\mathbf{c}')|\mathbf{c}\in V_m, \mathbf{c}'\in V_n\}$ is the set of edges between within-video clusters across videos. 
The edge weight $w(\mathbf{c},\mathbf{c}')$ is the distance between centroids of two within-video clusters $\mathbf{c},\mathbf{c}'$.
The solution to the $N$-dimensional assignment is a partition by dividing the graph $G$ into $K$ cliques $Z_1,Z_2,...,Z_K$. 
A clique $Z_k$, which is a subset of $N$ vertices from $N$ different vertex sets, defines the $k$-th global cluster. 
The induced sub-graphs of the cliques $Z_1,Z_2,...,Z_K$ are complete and disjoint. We denote the edge set of the induced sub-graph of $Z_k$ as $E_{Z_k}$. The cost of a clique is the sum of pairwise edge weights between the contained vertices. The cost of an assignment solution is the sum of the costs of all the $K$ cliques, \ie,
\begin{equation}
    \mathcal{L}(Z_1, Z_2, ..., Z_K)=\sum^K_{k=1}\sum_{(\mathbf{c},\mathbf{c}')\in E_{Z_k}} w(\mathbf{c},\mathbf{c}').
\end{equation}

\begin{figure}[t]
\centering
\includegraphics[width=0.7\linewidth]{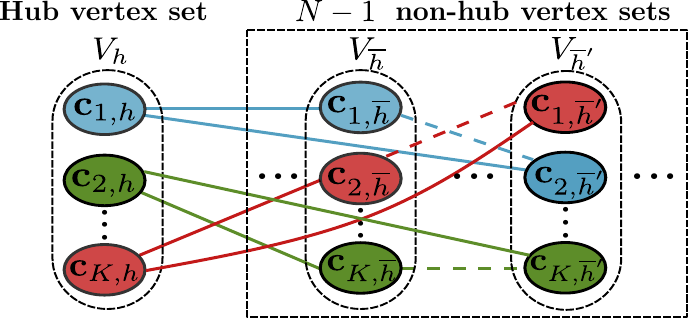}
\caption{In the $h$-th iteration, $h\in\{1,...,N\}$, the hub vertex set $V_h$ is chosen and an assignment is computed by assigning the vertices between $V_h$ and the $N-1$ non-hub vertex sets. Solid lines denote bipartite matching results between $V_h$ and non-hub vertex sets (step (1)). Dashed lines denote connections between non-hub vertex sets (step (2)).}
\label{fig:md_assign}
\end{figure}

In order to solve this NP-hard problem, we employ an iterative multiple-hub heuristic \cite{bandelt1994approximation}.
In each iteration, we choose a hub vertex set $V_h=\{\mathbf{c}_{k,h}|k=1,..,K\}$ and there are $(N-1)$ non-hub vertex sets.
We compute an assignment solution in each iteration in two steps, as is shown in Fig.~\ref{fig:md_assign}: (1) We first perform $(N-1)$ bipartite matchings between $V_h$ and each of the remaining non-hub vertex sets $V_{\overline h}$.
(2) Secondly, we determine the edge connection between pairs of non-hub vertex sets. On two non-hub vertex sets $V_{\overline h}, V_{\overline h'}$, we connect two vertices $\mathbf{c}_{i,\overline h} \in V_{\overline h}$ and $\mathbf{c}_{i',\overline h'} \in V_{\overline h'}$, if $\mathbf{c}_{i,\overline h}$ and $\mathbf{c}_{i',\overline h'}$ are connected to the same vertex $\mathbf{c}_{k,h}$ on $V_h$. 


After the two steps, every hub vertex $\mathbf{c}_{k,h}$, with $k\in\{1,..,K\}$ and all the non-hub vertices connected to $\mathbf{c}_{k,h}$ form the $k$-th clique $Z_k$. Therefore, the $N$-partite graph $G$ is partitioned into $K$ complete and disjoint subgraphs. By iterating over all possible initial hub vertex sets $h\in\{1,...,N\}$, we choose the assignment solution $f_{\hat h}$ which minimizes the assignment cost
\begin{equation}
    f_{\hat{h}} = \underset{h\in\{1,...,N\}}{\argmin} \sum_{(\mathbf{c},\mathbf{c}')\in E}f_h(\mathbf{c},\mathbf{c}')\cdot w(\mathbf{c},\mathbf{c}'),
\end{equation}
where $f_h(\mathbf{c},\mathbf{c}'), \forall (\mathbf{c}, \mathbf{c}')\in E$ is a binary indicator function that describes the edge connection: $f_h(\mathbf{c},\mathbf{c}')$ equals 1 when two vertices $\mathbf{c}, \mathbf{c}'$ are connected. 
The assignment solution $f_{\hat h}$ describes the partition which leads to the $K$ global clusters.


\subsection{Frame Labeling by Viterbi Decoding}\label{sec:frame_labeling}
Given the embedded feature sequence 
$\mathbf{e}_{1\sim T_n,n}$
of video $n$, we determine the optimal label sequence $\hat{y}_{1\sim T_n, n}$. The posterior probability can be factorized into the product of likelihoods and the probability of a given temporal order, \emph{i.e.}, $\hat{y}_{1\sim T_n, n}  = \underset{y_{1\sim T_n, n}}{\argmax} \ p(y_{1\sim T_n,n}|\mathbf{e}_{1\sim T_n,n})
     = \underset{y_{1\sim T_n, n}}{\argmax} \ \{\Pi^{T_n}_{i=1} p_n(\mathbf{e}_{i,n}|y_{i,n}) \cdot \Pi^{T_n}_{i=1} p_n(y_{i,n}|y_{1\sim i-1,n})\}$.
We fit a Gaussian model on each global cluster and compute the frame-wise likelihoods, \emph{i.e.}, $p_n(\mathbf{x}|k) = \mathcal{N}_k(\mathbf{x}; \mu_k, \Sigma_k), k\in\{1,...,K\}$. The temporal order constraint is used to limit the search space for the optimal label sequence by filtering out the sequences that do not follow the temporal order. 

The related works \cite{kukleva19, vidalmata2020joint} apply K-means on the frames of all the videos.
From the unified clustering they derive only a single temporal order of clusters for all the videos. 
However, this is an unrealistic assumption due to interchangeable steps in the activities, \eg, \textit{pour milk} and \textit{pour sugar} in \textit{making coffee}.
Instead, we can easily derive the temporal order for each video separately. We do so by sorting the within-video clusters according to the average timestamp of frames in each cluster.
The output of the Viterbi decoding is the optimal cluster label sequence $\hat{y}_{1\sim T_n, n}$.
More details are given in the supplementary.

\section{Experiments}
\subsection{Datasets \& Evaluation Metrics}\label{sec:datasets_and_eval_metrics}
We evaluate on Breakfast \cite{kuehne2014language}, the YouTube Instructions dataset (YTI) \cite{alayrac2016unsupervised} and 50 Salads \cite{stein2013combining}. Breakfast is comprised of 1712 videos recorded in various kitchens. There are 10 composite activities of breakfast preparation.
YTI is composed of 150 videos of 5 activities collected from YouTube. 50 Salads contains 50 videos of people preparing salads.
Following \cite{sener18, kukleva19, vidalmata2020joint}, we use the dense trajectory Fisher vector features (DTFV)~\cite{wang2013action} for Breakfast and 50 Salads, and features provided by Alayrac \etal~\cite{alayrac2016unsupervised} on YTI. 
We use the evaluation protocol in \cite{kukleva19} and report the performance in three metrics: (1)~Mean over Frames (MoF) is the frame-level accuracy over the frames of all the videos. More frequent or longer action instances have a higher impact on the result. (2)~Class-wise mean Intersection over Union (cIoU) is the average over the IoU performance for each class and penalizes segmentation results with dominating segments. (3)~The F1-score penalizes results with oversegmentation.
\subsection{Implementation Details}\label{sec:implementation_details}
For our SSTEN, we adapt the number of dilated residual layers $Q$ according to the dataset size: We set $Q=5$ for YTI (15k frames per activity subset on average) and $Q=10$ for Breakfast (360k) and 50 Salads (577k). The dimension of the hidden representation is set to 32. 
We set $\lambda$ in Eq.~\eqref{eq:recontr_loss_and_reltime_loss} to 0.002 (Breakfast), 0.01 (YTI) and 0.005 (50 Salads).
For clustering, we follow the protocol of \cite{sener18, alayrac2016unsupervised} and define the number of clusters $K$ separately for each activity as the maximum number of ground truth classes. The values of $K$ for the three datasets are provided in the supplementary material.

\subsection{Comparison with the State-of-the-Art}\label{sec:comparison_w_sota}
We compare with unsupervised learning methods, as well as weakly and fully supervised approaches on Breakfast (Table~\ref{tab:bf_state_of_the_art}), YTI (Table~\ref{tab:yti_state_of_the_art}) and 50 Salads (Table~\ref{tab:fs_state_of_the_art}).
Most unsupervised segmentation approaches yield cluster-aware segments that are aligned across all the videos~\cite{sener18, kukleva19, vidalmata2020joint, alayrac2016unsupervised, li2021action}. These approaches are evaluated with the \textbf{global Hungarian matching on all videos}, where the mapping between ground truth classes and clusters is performed on all the videos of an activity, which results in one mapping for each activity. The number of clusters $K$ is set to the maximum number of ground truth classes for each activity (\ie, \textbf{$K$=max.\#gt}). We focus on the performance comparison in this setting and follow this setting in all the ablation studies.

Two recent approaches perform clustering (\ie, TWFINTCH~\cite{sarfraz2021temporally}) or category-agnostic boundary detection (\ie, LSTM+AL~\cite{aakur19}) on each video individually, without solving the alignment among different clusters or segments across videos. For a fair comparison, these are evaluated by \textbf{local Hungarian matching on individual videos}, where a per-video best ground-truth-to-cluster-label mapping is determined using the ground truth on each video separately. This results in a separate label mapping for each video.
Following \cite{sarfraz2021temporally}, we also report results with $K$ set to the average number of actions for each activity (\ie, \textbf{$K$=avg.\#gt}) for a complete comparison.

In Table~\ref{tab:bf_state_of_the_art}, TAEC achieves strong results in comparison to the unsupervised state-of-the-art and is even comparable to weakly supervised approaches. 
Although approaches without solving the alignment of clusters across videos inherently lead to better scores in the evaluation settings of the local Hungarian matching, our approach still compares favorably. 

\begin{figure*}[!htb]
\centering
\includegraphics[width=0.9\linewidth]{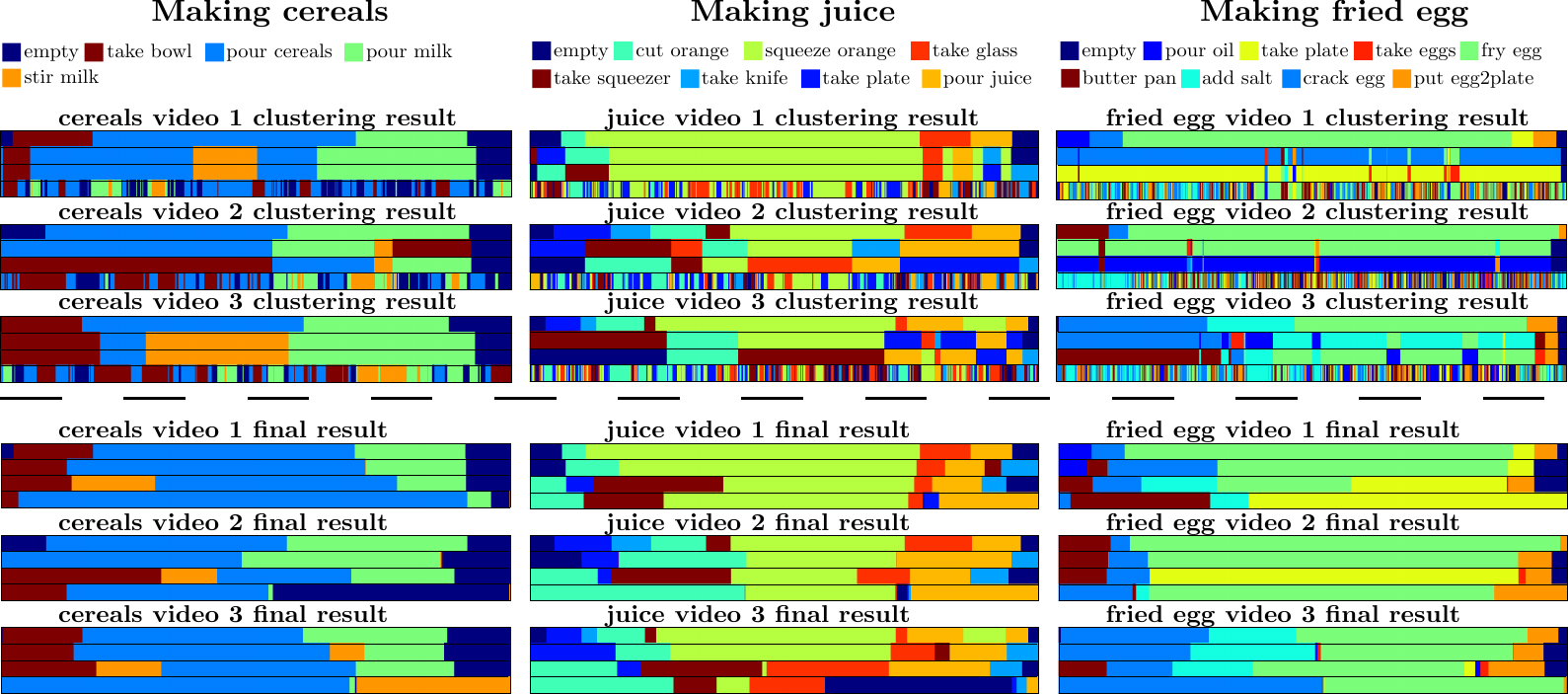}
\caption{Qualitative results of clustering and final segmentation (with global Hungarian matching) on 3 activities (3 videos each) on Breakfast. For each video, the 4-row-group displays the ground truth (1st row), TAEC (2nd row), TAEC with na\"ive assignment (3rd row, quantitative comparison in Sec.~\ref{sec:impact_cluster_assignment}), MLP+kmeans~\cite{kukleva19} (4th row). More quantitative and qualitative segmentation results are given in the supplementary.} 
\label{fig:segmentation_visualization_bf}
\end{figure*}

We compare qualitative results (with global Hungarian matching) of TAEC and MLP+kmeans \cite{kukleva19} on 3 Breakfast activities in Fig.~\ref{fig:segmentation_visualization_bf}. 
We see that our two-step clustering (the 2nd rows in all \textit{clustering result} plots) already leads to temporally consistent segments with relatively accurate boundaries of action instances, while K-means (the 4th rows in all \textit{clustering result} plots) results in serious oversegmentation. The Viterbi decoding further improves the segmentation by suppressing the oversegmentation and domination of incorrect clusters (the 2nd rows in all \textit{final result} plots). Moreover, MLP+kmeans \cite{kukleva19} follows the constraint of the fixed temporal order of segments on videos of each activity (the 4th rows in all \textit{final result} plots). In contrast, TAEC yields an individual temporal order for each video (the 2nd rows in all \textit{final result} plots). 
Additional qualitative results and evaluation scores are included in the supplemental material.

For the YouTube Instructions dataset, we follow the protocol of \cite{sener18, kukleva19, alayrac2016unsupervised} and report the results with and without considering background frames. Here, our TAEC outperforms all recent works in almost all of the metrics under all three settings.

50 Salads is a particularly challenging dataset for unsupervised approaches, as each video has a different order of actions and additionally includes many repetitive action instances. In the \textit{eval}-level of 12 classes, TAEC outperforms all approaches under the global Hungarian matching evaluation and achieves competitive results under the local Hungarian matching.
In the challenging \textit{mid}-level evaluation of 19 classes, the sequential nature of frames is less advantageous. Therefore, MLP+kmeans \cite{kukleva19} outperforms TAEC. Generally, in the local matching case, approaches without alignment across videos compare favorably.

\begin{table}[!htb]
\scriptsize
\centering
\caption{Comparison with state-of-the-art approaches on Breakfast (in \%). * denotes approach without segment alignment across videos, $\ddagger$ denotes our reimplementation, underlined scores are acquired from the author.}
\begin{tabular}{M{2cm}M{0.7cm}M{0.5cm}M{0.5cm}M{0.5cm}}
\toprule
\multicolumn{5}{c}{Breakfast} \\
\midrule
Approach & Supervision & MoF & IoU & F1\\
\midrule
MSTCN++~\cite{li2020ms} & full & 67.6 & - & - \\
G-FRNet~\cite{wang2020gated} & full & 67.7 & - & - \\
DTGRM~\cite{wang2021temporal} & full & 68.3 & - & - \\
SSTDA \cite{chen2020action} & full & 70.3 & - & - \\
BCN \cite{wang20boundary} & full & 70.4 & - & - \\
Global2local~\cite{gao2021global2local} & full & 70.7 & - & - \\
ASFormer~\cite{yi2021asformer} & full & \textbf{73.5} & - & - \\
\midrule
NN-Viterbi~\cite{richard2018neuralnetwork} & weak & 43.0 & - & - \\
D3TW~\cite{chang2019d3tw} & weak & 45.7 & - & - \\
TASL~\cite{lu2021weakly} & weak & 47.8 & - & - \\
CDFL~\cite{li19weakly} & weak & \textbf{50.2} & - & - \\
\midrule
\multicolumn{5}{c}{\textbf{Global Hungarian matching on all videos ($K$= max. \#gt)}}\\
\midrule
Mallow~\cite{sener18} & w/o & 34.6 & - & - \\
UNet+MLP~\cite{vidalmata2020joint} & w/o & 48.1 & - & \underline{29.9} \\
ASAL~\cite{li2021action} & w/o & \textbf{52.5} & - & \textbf{37.9} \\
MLP+kmeans~\cite{kukleva19} & w/o & 41.8 & - & 26.4  \\
MLP+kmeans~$\ddagger$ & w/o & 42.9 & 13.1 & 25.5\\
\textbf{TAEC} & w/o & 50.3 & \textbf{19.0} & 33.6\\
\midrule
\multicolumn{5}{c}{Local Hungarian matching on each video ($K$= \textbf{max.} \#gt)}\\
\midrule
LSTM+AL~\cite{aakur19}* & w/o & 42.9* & \textbf{46.9}* & - \\
UNet+MLP~\cite{vidalmata2020joint} & w/o & 52.2 & - & - \\
TWFINTCH~\cite{sarfraz2021temporally}* & w/o & 57.8* & - & - \\
MLP+kmeans~$\ddagger$ & w/o & 61.2 & 30.3 & 35.9\\
\textbf{TAEC} & w/o & \textbf{64.3} & 41.2 & \textbf{42.6}\\
\midrule
\multicolumn{5}{c}{Local Hungarian matching on each video ($K$= \textbf{avg.} \#gt)}\\
\midrule
TWFINCH~\cite{sarfraz2021temporally}* & w/o & \textbf{62.7}* & \textbf{42.3}* & - \\
MLP+kmeans~$\ddagger$ & w/o & 60.6 & 27.8 & 46.4\\
\textbf{TAEC} & w/o & 62.6 & 32.3 & \textbf{49.6}\\
\bottomrule
\end{tabular}
\label{tab:bf_state_of_the_art}
\end{table}

\begin{table}[!htbp]
\scriptsize
\centering
\caption{Comparison with the state-of-the-art on YTI (in \%). * denotes approach without segment alignment across videos, $\ddagger$ denotes our reimplementation, underlined scores are acquired from the author.}
\begin{tabular}{M{1.5cm}M{0.5cm}M{0.4cm}M{0.4cm}M{0.5cm}M{0.4cm}M{0.4cm}}
\toprule
\multicolumn{7}{c}{YouTube Instructions} \\
\midrule
Approach & Super-vision & MoF w/o bg & IoU w/o bg & F1 w/o bkg & MoF w bg & IoU w bg\\
\midrule
\multicolumn{7}{c}{\textbf{Global Hungarian matching on all videos ($K$= max. \#gt)}}\\
\midrule
Frank-Wolfe~\cite{alayrac2016unsupervised} & w/o & - & - & 24.4 & - & - \\
Mallow~\cite{sener18} & w/o & 27.8 & - & 27.0 & - & - \\
UNet+MLP~\cite{vidalmata2020joint} & w/o & \underline{28.9} & \underline{8.3} & 29.9 & - & -  \\
ASAL~\cite{li2021action} & w/o & 44.9 & - & \textbf{32.1} & - & - \\
MLP+kmeans~\cite{kukleva19} & w/o & 39.0 & 9.8 & 28.3 & 14.5 & 9.6  \\
MLP+kmeans~$\ddagger$ & w/o & 39.4 & 9.9 & 29.6 & 14.4 & 9.7  \\
\textbf{TAEC} & w/o & \textbf{46.6} & \textbf{10.7} & 29.5 & \textbf{17.0} & \textbf{10.5}   \\
\midrule
\multicolumn{7}{c}{Local Hungarian matching on each video ($K$= \textbf{max.} \#gt)}\\
\midrule
LSTM+AL~\cite{aakur19}* & w/o & - & - & 39.7* & - & - \\
MLP+kmeans~$\ddagger$ & w/o & 62.2 & 21.6 & 47.0 & 22.7 & 21.6  \\
\textbf{TAEC} & w/o & \textbf{67.9} & \textbf{23.7} & \textbf{49.4} & \textbf{24.8} & \textbf{23.6}  \\
\midrule
\multicolumn{7}{c}{Local Hungarian matching on each video ($K$= \textbf{avg.} \#gt)}\\
\midrule
TWFINCH~\cite{sarfraz2021temporally}* & w/o & 56.7* & - & 48.2* & - & - \\
MLP+kmeans~$\ddagger$ & w/o & 63.6 & 20.5 & \textbf{52.3} & 23.2 & 20.5  \\
\textbf{TAEC} & w/o & \textbf{65.3} & \textbf{20.9} & 51.0 & \textbf{23.9} & \textbf{20.8}  \\
\bottomrule
\end{tabular}
\label{tab:yti_state_of_the_art}
\end{table}

\begin{table}[!htb]
\scriptsize
\centering
\caption{Comparison with the state-of-the-art on 50 Salads (in \%). * denotes approach without segment alignment across videos, $\ddagger$ denotes our reimplementation.}
\begin{tabular}{cccM{0.4cm}M{0.4cm}M{0.4cm}}
\toprule
\multicolumn{6}{c}{50 Salads} \\
\midrule
Level & Approach & Supervision &  MoF & IoU & F1 \\
\midrule
\multirow{16}*{eval} & STCNN~\cite{lea2016segmental} & full & 72.0 & - & - \\
~ & EDTCN~\cite{lea2017temporal} & full &  73.4 & - & - \\
~ & MA~\cite{fermuller2018} & full & \textbf{88.5} & - & - \\
\cmidrule{2-6}
~ & \multicolumn{5}{c}{\textbf{Global Hungarian matching on all videos ($K$= max. \#gt)}}\\
\cmidrule{2-6}
~ & UNet+MLP~\cite{vidalmata2020joint} & w/o & 30.6 & - & -  \\
~ & ASAL~\cite{li2021action} & w/o & 39.2 & - & - \\
~ & MLP+kmeans~\cite{kukleva19} & w/o & 35.5 & - & -  \\
~ & MLP+kmeans~$\ddagger$ & w/o & 37.9 & 24.6 & 40.2 \\
~ & \textbf{TAEC} & w/o & \textbf{48.4} & \textbf{26.0} & \textbf{44.8} \\
\cmidrule{2-6}
~ & \multicolumn{5}{c}{Local Hungarian matching on each video ($K$= \textbf{max.} \#gt)}\\
\cmidrule{2-6}
~ & LSTM+AL~\cite{aakur19}* & w/o & \textbf{60.6}* & - & - \\
~ & MLP+kmeans~$\ddagger$ & w/o & 58.0 & 33.7 & 49.8 \\
~ & \textbf{TAEC} & w/o & 59.7 & \textbf{35.0} & \textbf{54.4} \\
\cmidrule{2-6}
~ & \multicolumn{5}{c}{Local Hungarian matching on each video ($K$= \textbf{avg.} \#gt)}\\
\cmidrule{2-6}
~ & TWFINCH~\cite{sarfraz2021temporally}* & w/o & \textbf{71.1}* & - & -\\
~ & MLP+kmeans~$\ddagger$ & w/o & 51.5 & 22.3 & 43.4 \\
~ & \textbf{TAEC} & w/o & 59.7 & \textbf{35.7} & \textbf{54.7} \\
\midrule
\multirow{20}*{mid} & SSTDA~\cite{chen2020action} & full &  83.2 & - & - \\
~ & MSTCN++~\cite{li2020ms} & full & 83.7 & - & - \\
~ & HASR~\cite{ahn2021refining} & full & 83.9 & - & - \\
~ & ASRF~\cite{ishikawa2021alleviating} & full & 84.5 & - & - \\
~ & ASFormer~\cite{yi2021asformer} & full & \textbf{85.6} & - & - \\
\cmidrule{2-6}
~ & NNViterbi~\cite{richard2018neuralnetwork} & weak & 49.4 & - & - \\
~ & CDFL~\cite{liu19} & weak & \textbf{54.7} & - & - \\
\cmidrule{2-6}
~ & \multicolumn{5}{c}{\textbf{Global Hungarian matching on all videos ($K$= max. \#gt)}}\\
\cmidrule{2-6}
~ & UNet+MLP~\cite{vidalmata2020joint} & w/o & 24.2 & - & -  \\
~ & ASAL~\cite{li2021action} & w/o & \textbf{34.4} & - & - \\
~ & MLP+kmeans~\cite{kukleva19} & w/o & 30.2 & - & -  \\
~ & MLP+kmeans~$\ddagger$ & w/o & 29.1 & \textbf{15.7} & \textbf{23.4} \\
~ & \textbf{TAEC} & w/o & 26.6 & 14.9 & \textbf{23.4} \\
\cmidrule{2-6}
~ & \multicolumn{5}{c}{Local Hungarian matching on each video ($K$= \textbf{max.} \#gt)}\\
\cmidrule{2-6}
~ & MLP+kmeans~$\ddagger$ & w/o & \textbf{55.6} & \textbf{29.6} & 39.6 \\
~ & \textbf{TAEC} & w/o & 50.2 & 29.4 & \textbf{40.3} \\
\cmidrule{2-6}
\cmidrule{2-6}
~ & \multicolumn{5}{c}{Local Hungarian matching on each video ($K$= \textbf{avg.} \#gt)}\\
\cmidrule{2-6}
~ & TWFINCH~\cite{sarfraz2021temporally}* & w/o & \textbf{66.5}* & - & -\\
~ & MLP+kmeans~$\ddagger$ & w/o & 53.4 & 28.1 & 39.0 \\
~ & \textbf{TAEC} & w/o & 51.9 & \textbf{30.2} & \textbf{41.9} \\
\bottomrule
\end{tabular}
\label{tab:fs_state_of_the_art}
\end{table}

\subsection{Embedding and Clustering}\label{sec:embedding_and_clustering}
We first evaluate our SSTEN embedding in combination with K-means and two-step clustering on three feature types: the AlexNet fc6 features \cite{krizhevsky2012imagenet} pre-trained on ImageNet \cite{deng2009imagenet}, I3D features \cite{carreira2017quo} pre-trained on the Kinetics dataset \cite{kay2017kinetics}, and the precomputed dense trajectory Fisher vectors (DTFV) \cite{wang2013action}. 
We also report results of raw features without any temporal embedding. 
For a fair comparison, we reduce the dimensions of the three features without embedding to 32 via PCA. We conduct the experiments on Breakfast and report the results with and without our SSTEN embedding in Table~\ref{tab:comparison_w_wo_SSTEN}. 


\begin{table}[!htb]
\scriptsize
\centering
\caption{Comparison of features of SSTEN embedding, together with clustering methods on Breakfast (in \%).}
\begin{tabular}{ M{0.5cm}M{0.8cm}M{0.4cm}M{0.4cm}M{0.4cm}M{0.4cm}M{0.4cm}M{0.4cm}}
\toprule
\multicolumn{2}{c}{Model}  & \multicolumn{3}{c}{K-means} & \multicolumn{3}{c}{Two-step clustering} \\
\cmidrule(r){1-2}\cmidrule(r){3-5}\cmidrule{6-8}
Feature & Embedding & MoF & IoU & F1 & MoF & IoU & F1 \\
\midrule
AlexNet &\multirow{3}*{w/o}  & 25.9 & 11.3 & 22.3 & 33.7 & 10.7 & 20.2 \\
 I3D & ~ & \textbf{33.4} & \textbf{14.4} & \textbf{25.6} & \textbf{37.7} & \textbf{14.3} & \textbf{26.1} \\
DTFV & ~  & 30.8 & 11.8 & 23.0 & 34.5 & 12.1 & 22.0 \\
\midrule
AlexNet & \multirow{3}*{SSTEN}  & 33.0 & 14.2 & 27.0 & 39.1 & 16.1 & 30.7 \\
I3D & ~ &  37.9 & \textbf{18.7} & \textbf{33.3} & 45.2 & \textbf{20.5} & \textbf{35.1} \\
DTFV & ~ & \textbf{39.3}  & 17.8 & 31.9 & \textbf{50.3} & 19.0 & 33.6  \\
\bottomrule
\end{tabular}
\label{tab:comparison_w_wo_SSTEN}
\end{table}

\textbf{Comparison of raw features without embedding.} 
Among the three types of features without temporal embedding, I3D achieves the best performance, while AlexNet features lead to the worst results. 
AlexNet features are computed from individual spatial frames. 
On the contrary, each frame feature of DTFV and I3D is computed based on a chunk of its temporal neighbor frames. Therefore, the features already carry intrinsic temporal consistency. 
Furthermore, the two-stream I3D model can leverage both RGB and optical flow.
Therefore, I3D features achieve a better performance than DTFV, which rely on handcrafted dense trajectories.

\textbf{Comparison of SSTEN embeddings learned on different features.} When comparing the SSTEN embeddings to the performance of the raw features, we see that SSTEN leads to a significant performance gain for both clustering methods. For DTFV, the performance improvements by SSTEN are MoF 8.5\%, IoU 6.0\%, F1 8.9\% with K-means and MoF 15.8\%, IoU 6.9\%, F1 11.6\% with two-step clustering. 

Among the three types of SSTEN embedded features, I3D has slightly better IoU and F1 scores while
DTFV leads to the best MoF scores for both, K-means and the two-step clustering. Overall, the SSTEN embeddings learned from these two features perform comparably. We conduct the following experiments using DTFV, which is also used in related works.

\subsection{Impact of Loss Terms on Clustering}\label{sec:impact_of_loss_terms_on_clustering}
\begin{figure}[t]
\centering
\includegraphics[width=\linewidth]{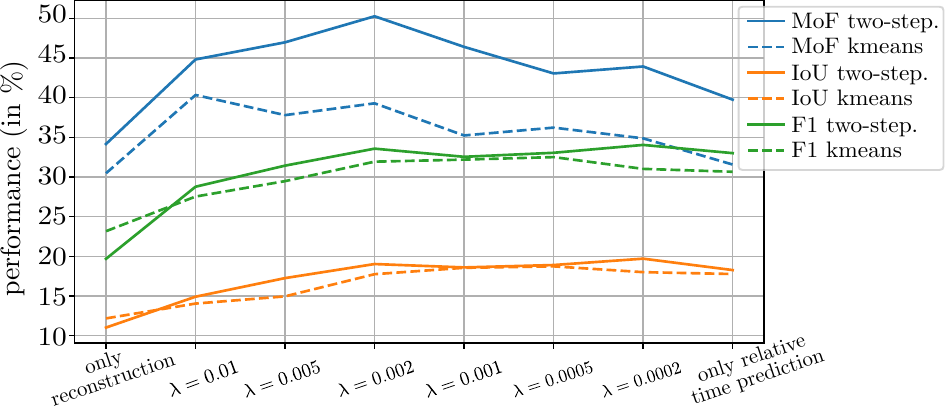}
\caption{Segmentation performance of both clustering methods on SSTEN embeddings with different $\lambda$ on Breakfast.}
\label{fig:plot_lambda_vs_result_bf}
\end{figure}

To evaluate the impact of the two loss terms in Eq.~\eqref{eq:recontr_loss_and_reltime_loss}, we plot the quantitative segmentation results of SSTEN with both K-means and the two-step clustering w.r.t. different reconstruction loss coefficients $\lambda$ in Fig.~\ref{fig:plot_lambda_vs_result_bf}.
In general, two-step clustering leads to a better performance than K-means for almost all $\lambda$ values (except for the case of only reconstruction loss). With decreasing $\lambda$, the relative time prediction loss has an increasing impact and the embedded features have better global temporal consistency, which explains the increasing IoU and F1 scores. However, at extremely small $\lambda$ values, the embedded features overfit to the relative time prediction task, which results in saturated IoU and F1 scores, and a significant drop in MoF for both K-means and two-step clustering.

\begin{figure}[!htb]
\centering
\includegraphics[width=0.8\linewidth]{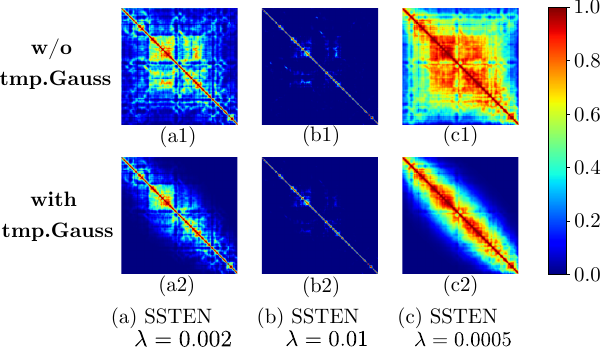}
\caption{Frame-to-frame similarity matrices of SSTEN embeddings for the same Breakfast video. 
Columns show the similarity matrices for different $\lambda$, while the rows show results without (top) and with (bottom) temporal Gaussian kernel.}
\label{fig:affinity_lambda}
\end{figure}

To intuitively illustrate the loss term impact on the two-step clustering, we plot the similarity matrices for SSTEN embeddings trained with three different $\lambda$ in Fig.~\ref{fig:affinity_lambda}. Here, we look at the similarity matrices with temporal Gaussian kernel (bottom row). Intuitively, the similarity matrix with clear diagonal block structure (Fig.~\ref{fig:affinity_lambda}(a2)), which is the result of an appropriate ratio between the reconstruction loss and relative time prediction loss ($\lambda=0.002$), leads to the best segmentation performance. When $\lambda$ becomes larger (\eg, $\lambda=0.01$), the reconstruction loss has a larger impact and the diagonal block structure (Fig.~\ref{fig:affinity_lambda}(b2)) becomes pale. Therefore, the performances of embedded features with $\lambda=0.005$, $\lambda=0.01$ and only reconstruction loss degrade successively. On the other hand, for extremely small $\lambda$ values (\eg, $\lambda=0.0005$), the block diagonal structure (Fig.~\ref{fig:affinity_lambda}(c2)) becomes noisy due to overfitting on relative time prediction.

Therefore, both the reconstruction and the relative timestamp prediction loss, when combined in an appropriate ratio, are indispensable to learn the effective representation that preserves both spatial layout and the temporal information.

\subsection{Impact of Cluster Assignment}\label{sec:impact_cluster_assignment}
In this ablation study, we evaluate the efficacy of the global cluster assignment. For two-step clustering, we evaluate two strategies of grouping within-video clusters into global clusters: (1) the na\"ive assignment, for which we order the sub-clusters according to the average timestamp and simply group the $k$-th sub-clusters of all videos into a global cluster, \ie, the global cluster $Z_k=\{\mathbf{c}_{k,n}|n=1,..,N\}$, and (2) the global cluster assignment, as detailed in Sec.~\ref{sec:global_cluster_assignment}. 

In order to show how the different cluster assignment strategies affect the clustering result, we report both, the results of the two-step clustering (before Viterbi decoding) and the final segmentation performance (after Viterbi decoding) on Breakfast and 50 Salads in Table~\ref{tab:impact_global_cluster_assignment}.
%
\begin{table}[!htb]
\centering
\scriptsize
\caption{Impact of cluster assignment strategies for two-step clustering on SSTEN embeddings on Breakfast and 50 Salads (\emph{eval} level, \ie, 12 classes) (in \%).}
\begin{tabular}{M{0.5cm}M{1.2cm}M{0.3cm}M{0.3cm}M{0.3cm}M{0.3cm}M{0.3cm}M{0.3cm}}
\toprule
\multirow{2}*{Dataset} &  \multirow{2}*{Strategy} &  \multicolumn{3}{c}{Clustering} &  \multicolumn{3}{c}{Final} \\
\cmidrule(r){3-5}\cmidrule{6-8}
~ &  ~ &  MoF &  IoU &  F1 &  MoF &  IoU &  F1 \\
\midrule
\multirow{2}*{Breakfast} &  global cluster & \textbf{38.6} & \textbf{13.7} & \textbf{25.9} & \textbf{50.3} &  \textbf{19.0} &  \textbf{33.6} \\ 
~ &  na\"ive & 25.1 & 12.4 & 23.9 & 42.3 &  17.7 &  32.0  \\
\midrule
\multirow{2}*{50 Salads} &  global cluster &  \textbf{45.3}  &  \textbf{23.0} &  \textbf{43.7} &  \textbf{48.4} &  \textbf{26.0} &  \textbf{44.8} \\ 
~ &  na\"ive &  28.9 &  16.2 &  32.9 &  35.0 &  22.7 &  38.0  \\
\bottomrule
\end{tabular}
\label{tab:impact_global_cluster_assignment}
\end{table}
The global cluster assignment outperforms the na\"ive assignment by a large margin for both, clustering results and the final segmentation results, on both datasets. The advantage of the global cluster assignment is even more evident on 50 Salads. 

We illustrate exemplary qualitative results of the clustering and the final segmentation for 3 activities (with 3 videos each) on Breakfast in Fig.~\ref{fig:segmentation_visualization_bf}.
For each video, the 4-row group displays the ground truth (1st row), the result with global cluster assignment (2nd row) and the result with na\"ive assignment (3rd row). 
The 4th row shows the result of MLP+kmeans~\cite{kukleva19}. 
By comparing all the 3rd rows of \textit{cereals video [id] \underline{final} result} in Fig.~\ref{fig:segmentation_visualization_bf}, we see that the na\"ive assignment simply assumes the sub-clusters in the same temporal order in each video belong to the same global cluster, while they might not be close to each other in the feature space. On the contrary, the global cluster assignment (the 2nd rows of \textit{cereals video [id] \underline{final} result}) yields an optimal assignment solution with respect to the pairwise distances between sub-clusters, resulting in different orderings of sub-clusters on each video. Note that on some videos, global cluster assignment could lead to the same assignment result as na\"ive assignment.

\section{Conclusion}
We proposed a new pipeline for the unsupervised learning of action segmentation. 
For the feature embedding, we propose a temporal-aware embedding network that performs sequence-to-sequence learning with the pretext tasks of relative timestamp prediction and feature reconstruction. For clustering, we propose a two-step clustering schema, consisting of within-video clustering and cross-video global cluster assignment. The temporal embedding of sequence-to-sequence learning together with two-step clustering is proven to be a well-suitable combination that considers the sequential nature of frames in both processing steps. Ultimately, we combine the temporal embedding with a frame-to-cluster assignment based on Viterbi decoding, which achieves the unsupervised state-of-the-art on three challenging benchmarks.

\newpage

\appendix
\section*{Supplementary}
\section{Introduction}
For additional insights into TAEC, we introduce the background of spectral clustering in Sec.~\ref{sec:details_normalized_cut} and give details of the Viterbi decoding in Sec.~\ref{sec:frame_labeling}.
We perform more ablation studies on comparing baseline embeddings and clustering methods (Sec.~\ref{sec:embed_cluster}), scaling of spatio-temporal similarity (Sec.~\ref{sec:impact_scaling_in_spatiotmeporal_similarity}), cluster ordering (Sec.~\ref{sec:impact_cluster_order}), decoding strategies (Sec.~\ref{sec:impact_decoding_strategies}). 
Finally, we provide more quantitative (Sec.~\ref{sec:quantatitive_results}) and qualitative segmentation results (Sec.~\ref{sec:qualitative_results}) on the three datasets. 

\section{Method}
\subsection{Spectral Clustering}\label{sec:details_normalized_cut}
Background information related to Sec.~3.2.1 in the main manuscript: Given the embedded feature sequence $\mathbf{e}_1, \mathbf{e}_2,...,\mathbf{e}_T$, we build a frame-to-frame similarity graph $G\in \mathbb{R}^{T\times T}$, whose edge weight $g(i,j),\ i,j\in\{1,..., T\}$ represents the similarity between frame $i$ and frame $j$. 
Grouping the frames into $K$ clusters can be interpreted as a graph partition problem by cutting edges on $G$, resulting in $K$ subgraphs $G_1, G_2, ..., G_K$.
The normalized cut (Ncut) problem \cite{shi2000normalized} is employed to compute a balanced partition by minimizing the energy
\begin{equation}\label{eq:normalized_cut}
    \mathcal{L}_{cut}(G_1, G_2, ..., G_K) = \frac{1}{2}\sum_{k=1}^K \frac{W(G_k, \overline{G}_k)}{\operatorname{vol}(G_k)},
\end{equation}
where $W(G_k, \overline{G}_k)$ represents the sum of edge weights between elements in the subgraph $G_k$ and elements of all the other subgraphs, \emph{i.e.}, the sum of weights of edges to be cut. $\operatorname{vol}(G_k)$ is the sum of weights of edges within the resulting subgraph $G_k$. Spectral clustering \cite{vonluxburg07} is a relaxed solution to this NP-hard minimization problem in Eq.~\eqref{eq:normalized_cut} and has shown good performance on many graph-based clustering problems, \eg~\cite{huang2012affinity, li2015superpixel, liu2004segmentation}. Note that while K-means operates on Euclidean distance in the feature space and assumes convex and isotropic clusters, spectral clustering can find clusters with non-convex boundaries.

\subsection{Frame Labeling by Viterbi Decoding}\label{sec:frame_labeling}
Additional explanations to Sec.~3.3 in the main manuscript: The global cluster assignment delivers the ordered clusters on each video, which are aligned across all videos. To compute the final segmentation, we use the resulting ordering and decode each video into a sequence of $K$ temporally consistent segments. That is, we determine the optimal label sequence $\hat{y}_{1\sim T_n, n}=\{y_{1,n},..., y_{T_n,n}\}$ by re-assigning each frame to one of the temporally ordered clusters. 

Given the embedded feature sequence $\mathbf{e}_{1\sim T_n,n}= \{\mathbf{e}_{1,n},...,\mathbf{e}_{T_n, n} \}$ and the temporal order of the clusters, we search for the optimal label sequence that maximizes the probability $p(y_{1\sim T_n,n}|\mathbf{e}_{1\sim T_n,n})$. Following \cite{richard2018neuralnetwork}, this posterior probability can be factorized into the product of likelihoods and the probability of a given temporal order, \emph{i.e.}, 
\begin{align}\label{eq:posterior_prob}
     & \hat{y}_{1\sim T_n, n}  = \underset{y_{1\sim T_n, n}}{\argmax} \ p(y_{1\sim T_n,n}|\mathbf{e}_{1\sim T_n,n})\nonumber \\
     & = \underset{y_{1\sim T_n, n}}{\argmax} \ \{\Pi^{T_n}_{t=1} p(\mathbf{e}_{t,n}|y_{t,n}) \cdot \Pi^{T_n}_{t=1} p_n(y_{t,n}|y_{1\sim (t-1),n})\}\nonumber \\
     & = \underset{y_{1\sim T_n, n}}{\argmax} \ \{ \Pi^{T_n}_{t=1} p(\mathbf{e}_{t,n}|y_{t,n}) \cdot  p_n(y_{t,n}|y_{t-1,n}) \}
\end{align}

Here the likelihood $p(\mathbf{e}_{t,n}|y_{t,n})$ is the probability of a frame embedding $\mathbf{e}_{t,n}$ from the video $n$ belonging to a cluster. Therefore, we fit a Gaussian distribution on each global cluster and compute the frame-wise likelihoods with the Gaussian model, \ie,
\begin{equation}
    p(\mathbf{x}|k) = \mathcal{N}_k(\mathbf{x}; \mu_k, \Sigma_k), k\in\{1,...,K\}.
\end{equation}

$p_n(y_{t,n}|y_{t-1,n})$ is the transition probability from label $y_{t-1,n}$ at frame $t-1$ to label $y_{t,n}$ at frame $t$, which is defined by the temporal order of clusters. We denote the set of frame transitions defined by the temporal order of clusters on the $n$-th video by $O_n$, \eg, for the temporal order of $a\rightarrow b\rightarrow c\rightarrow d$, $O_n = \{a\rightarrow b, b\rightarrow c, c\rightarrow d\}$. 
The transition probability is binary, \emph{i.e.}, 
\begin{align}
    &p_n(y_{i,n}|y_{i-1, n})\\\notag
    &\quad=\mathbbm{1}(y_{i,n}=y_{i-1,n} \lor y_{i-1,n}\rightarrow y_{i,n} \in O_n).
\end{align}
This means that we allow either a transition to the next cluster according to the temporal order, or we keep the cluster assignment of the previous frame. 

Note that in \textbf{two-step clustering}, we derive the temporal order of clusters on each video separately, by sorting the clusters on the video according to the average timestamp. Therefore, we have an individual $O_n$ for each video $n$. On the contrary, in \textbf{K-means}, there is a uniform order of global clusters for all the videos and $O_n$ is thus the same for each video $n$.

The Viterbi algorithm for solving Eq.~\eqref{eq:posterior_prob} is performed in an iterative process using dynamic programming, \emph{i.e.},
\begin{align}
p(y_{1\sim t,n}|\mathbf{e}_{1\sim t,n})&=\\\notag
\underset{y_{t,n}}{\max} \ &\{p(y_{1\sim t-1, n} | \mathbf{e}_{1\sim t-1, n})\\\notag
&\cdot p(\mathbf{e}_{t,n}|y_{t,n}) \cdot p(y_{t,n}|y_{t-1,n})\}.
\end{align}
%
The sequences that do not follow the temporal order will be filtered out in an early stage to narrow down the search range for the optimal label sequence. 
The output of the Viterbi decoding is the optimal cluster label sequence, \ie,~$\hat{y}_{1\sim T_n, n}$.

\section{Additional Results}

\subsection{Embedding and clustering}\label{sec:embed_cluster}
\begin{figure}[t]
\centering
\includegraphics[width=\linewidth]{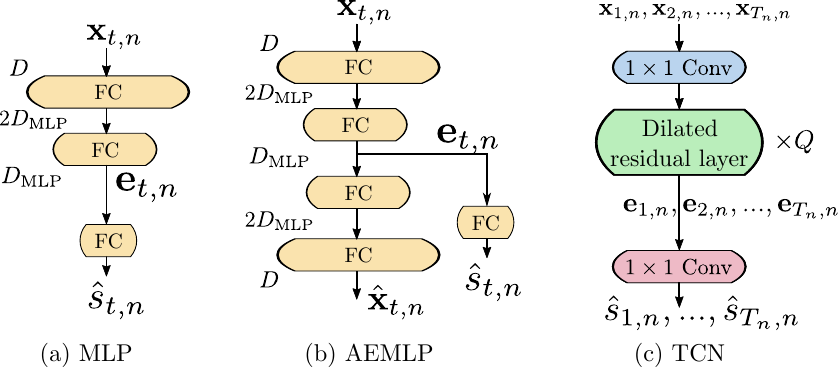}
\caption{Three baseline variants of embedding networks.
}
\label{fig:network_structure_baseline}
\end{figure}

Further, we compare our SSTEN embedding with three baseline variants (shown in Fig.~\ref{fig:network_structure_baseline}): MLP temporal embedding, autoencoder with MLP (AEMLP) and temporal convolutional network (TCN), in combination with the two clustering methods.  

\textit{MLP} uses three FC layers for relative timestamp prediction. \textit{AEMLP} uses MLP-based autencoder for both relative timestamp prediction and feature reconstruction. \textit{TCN} deploys $Q$ stacked dilated residual layers only for relative timestamp prediction.

Here, we also implement the Rankloss MLP embedding~\cite{sener18} for reference. We report the performance of these five embeddings in Table~\ref{tab:comparison_embeddings_clusterings}.

\begin{table}[!htb]
\scriptsize
\centering
\caption{Comparison of combinations of embeddings and clustering methods on Breakfast (in \%).}
\begin{tabular}{ M{0.5cm}M{1.2cm}M{0.3cm}M{0.3cm}M{0.3cm}M{0.3cm}M{0.3cm}M{0.3cm}}
\toprule
\multicolumn{2}{c}{Model}  & \multicolumn{3}{c}{K-means} & \multicolumn{3}{c}{Two-step clustering} \\
\cmidrule(r){1-2}\cmidrule(r){3-5}\cmidrule{6-8}
Feature & Embedding & MoF & IoU & F1 & MoF & IoU & F1 \\
\midrule
\multirow{5}*{DTFV} & Rankloss \cite{sener18} & 35.2 & 15.6 & 28.8 & 34.7 & 13.4 & 23.7 \\
~ & MLP & \textbf{42.9} & 13.1 & 25.5 & 32.7 & 10.9 & 21.2 \\
~ & AEMLP & 34.7 & 13.6 & 25.8 & 32.6 & 11.6 & 21.4 \\
~ & TCN & 33.4  & \textbf{17.8} & 31.3 & 40.4 & \textbf{19.1} & 32.9 \\
~ & SSTEN & 39.3  & \textbf{17.8} & \textbf{31.9} & \textbf{50.3} & 19.0 & \textbf{33.6} \\
\bottomrule
\end{tabular}
\label{tab:comparison_embeddings_clusterings}
\end{table}

\textbf{Comparison of the five embeddings.} We learn the five embeddings (Rankloss MLP, MLP, AEMLP, TCN and SSTEN) on the DTFV features. Here, the Rankloss MLP (consisting of two FC layers) is trained with a ranking loss. We use the initialization of uniform segmentation as the temporal prior to train the model with only one iteration.

TCN and SSTEN are both networks for sequence-to-sequence learning, while Rankloss MLP, MLP and AEMLP are trained on individual frames. By comparing the performance between these two groups in Table~\ref{tab:comparison_embeddings_clusterings}, we see that sequence-to-sequence learning leads to better performance, especially when combined with the two-step clustering, which results in clusters with better temporal consistency.

\begin{figure}[!t]
\centering
\includegraphics[width=\linewidth]{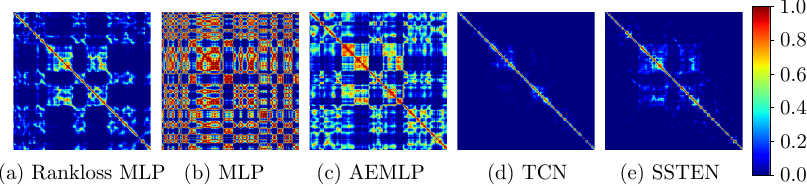}
\caption{Frame-to-frame similarity matrices of the embedded sequences of the same video (computed by different embedding networks from DTFV features) on Breakfast.}
\label{fig:affinity_embedding_networks}
\end{figure}

For the two-step clustering, we also plot the frame-to-frame similarity matrices (spatial Gaussian kernel) of the five embeddings for the same Breakfast video in Fig.~\ref{fig:affinity_embedding_networks}. The plots show that Rankloss MLP, MLP and AEMLP, which are trained on individual frames, do not expose an appropriate temporal structure. There are noisy block patterns even in positions far away from the diagonal, which results in noisy clusters and thus, leads to erroneous temporal orders and inferior assignment results in the two-step clustering. The least noisy Rankloss MLP has the highest performance among these three. On the contrary, TCN and SSTEN embedded features, which show a clear diagonal block structure in the similarity graph, achieve a better performance in the two-step clustering. 
This verifies that the sequence-to-sequence embedding learning (TCN and SSTEN) and two-step clustering are a well-suited combination to address the sequential nature of frames in both processing steps of feature embedding and clustering. 

Considering K-means clustering, the merit of having a better sequential nature of the embedded features via sequence-to-sequence learning can also be seen from the higher IoU and F1 scores (TCN: IoU 17.8\%/F1 31.3\%, \emph{vs.} SSTEN: 17.8\%/31.9\%), as these penalize dominating segments and oversegmentation. 

In contrast to TCN, SSTEN can preserve the spatial layout of the input features due to the feature reconstruction via the autoencoder.
By comparing TCN and SSTEN, we see that the SSTEN embedding with feature reconstruction leads to a boost in the MoF score. The marginal improvement of AEMLP over MLP is due to the fact that the MLP structure with only FC layers is not well-suited for feature reconstruction.

\textbf{Comparison between K-means and two-step clustering.} Considering the performance of the five embeddings with the two clustering methods, we see that K-means leads to higher scores on the inferior embeddings (Rankloss MLP, MLP and AEMLP) trained on individual frames, while two-step clustering performs better on sequence-to-sequence learning-based embeddings (TCN and SSTEN). When combined with the proposed SSTEN embedding, two-step clustering outperforms K-means by a large margin in terms of the MoF score. We also tried applying K-means on each video separately. However, the performance dropped significantly. K-means depends only on the spatial distance and results in oversegmentation, which leads to erroneous temporal order on each video and thus, an inferior global cluster assignment.

\subsection{Impact of Scaling in Spatiotemporal Similarity}\label{sec:impact_scaling_in_spatiotmeporal_similarity}
We perform spectral clustering with the proposed spatiotemporal similarity.
Here, we analyze the impact of the scaling factors in the spatial and temporal Gaussian kernels, \ie, $\sigma_{\text{spat}}^2$ and $\sigma_{\text{tmp}}^2$. These adjust the extent to which two frames are considered similar to each other and influence the clustering quality. The experiments are conducted for SSTEN embeddings on Breakfast.

\begin{figure}[t]
\centering
\includegraphics[width=0.8\linewidth]{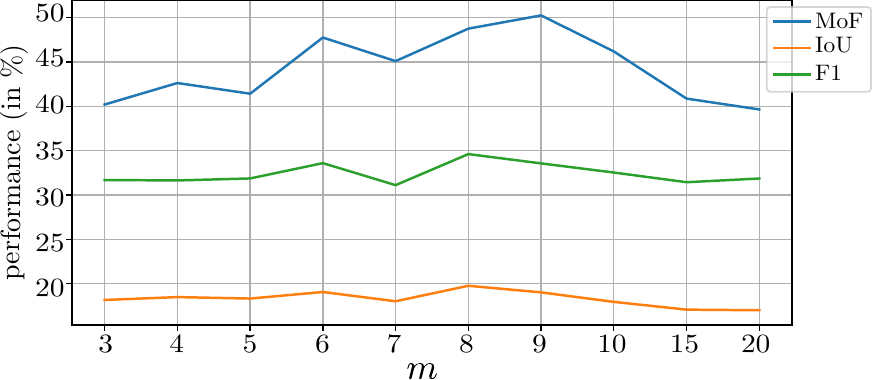}
\caption{Segmentation performance of the two-step clustering on SSTEN embeddings ($\lambda=0.002$) with different $m$ (for local scaling) on Breakfast.}
\label{fig:plot_m_vs_result}
\end{figure}

\textbf{Impact of the scaling of the spatial Gaussian kernel.} 
For local scaling, we set $\sigma_{\text{spat}}^2=\sigma_i\sigma_j$, where $\sigma_i$ is the distance from $\mathbf{e}_i$ to its $m$-th nearest neighbor in the feature space. 
The resulting segmentation performance w.r.t. $m$ is shown in Fig.~\ref{fig:plot_m_vs_result}. 
With $m$ varying in the range of 3 to 20, the IoU and F1 scores remain stable. 
There is a range of $m\in\{8, 9\}$ where the best MoF scores are achieved, whereas for other scaling parameters, the MoF score drops.
Thus, we set $m=9$ for all following evaluations. 

For comparison, we also set $\sigma_{\text{spat}}$ to fixed values (without local scaling) and report the segmentation performance
in Table~\ref{tab:sigma_spat_vs_result_bf}. 
We achieve great results at smaller $\sigma_{\text{spat}}$ values (0.5 and 0.7).

However, with increasing $\sigma_{\text{spat}}$ the MoF score drops significantly, while there are only minor fluctuations in IoU and F1. Apparently, $\sigma_{\text{spat}}$ has a large impact on the clustering quality. The local scaling eases the effort of tuning $\sigma_{\text{spat}}$ by dynamically determining the scaling factor.

\begin{table}[!htb]
\scriptsize
\centering
\caption{Segmentation performance of two-step clustering on SSTEN embeddings ($\lambda=0.002$) with respect to a fixed spatial scaling factor $\sigma_{\text{spat}}$ (without local scaling) on Breakfast (in \%).}
\begin{tabularx}{0.35\textwidth}{cXcXcXc}
\toprule
$\sigma_{\text{spat}}$ & & MoF & & IoU & & F1 \\
\midrule
 0.5 & & 47.1 & & 18.0 & & 33.4  \\
  0.7 & & \textbf{48.0} & & 18.5 & & \textbf{33.9}  \\
   1 & & 41.1 & & \textbf{19.2} & & 32.9  \\
    2 & & 38.4 & & 18.6 & & 32.2  \\
     10 & & 35.0 & & 17.9 & & 31.2  \\
\bottomrule
\end{tabularx}
\label{tab:sigma_spat_vs_result_bf}
\end{table}

\textbf{Impact of the scaling of the temporal Gaussian kernel.} 
The temporal Gaussian kernel is operated on the temporal distance between frames in a video. With $\sigma_{\text{tmp}}^2 = 2{\sigma'}^2$, the term $\exp{\left(-(s_i - s_j)^2 / (2\sigma'^2) \right)}$ is in the standard form of a Gaussian kernel. We set $\sigma'=1/6$ so that the $6\sigma'$ range of the temporal Gaussian is equal to the video length (since the length of each video is normalized to 1 for the relative timestamp prediction). The segmentation performance with respect to $\sigma'$ is shown in Table~\ref{tab:sigma_tmp_vs_result_bf}.
\begin{table}[!htb]
\scriptsize
\centering
\caption{Segmentation performance of two-step clustering on SSTEN embeddings ($\lambda=0.002$) with respect to the temporal scaling factor $\sigma'$ ($\sigma_{\text{tmp}}^2 = 2{\sigma'}^2$ on Breakfast (in \%).}
\begin{tabularx}{0.45\textwidth}{cXcXcXc}
\toprule
$\sigma'$ ($\sigma_{\text{tmp}}^2 = 2{\sigma'}^2$) & & MoF & & IoU & & F1 \\
\midrule
 $\infty$ (w/o tmp. Gauss) & &  41.5 & & 16.5 & & 30.6  \\
1/3 & & 43.5 & & 16.9 & & 31.3  \\
1/6 & & \textbf{50.3} & & \textbf{19.0} & & 33.6  \\
1/12 & & 44.3 & & 18.5 & & \textbf{34.1}  \\
\bottomrule
\end{tabularx}
\label{tab:sigma_tmp_vs_result_bf}
\end{table}
Apparently, $\sigma'=1/6$ leads to the best result. Here, we also evaluate the case without the temporal Gaussian kernel, which leads to a drop in performance. 
The impact of the temporal Gaussian kernel on similarity matrices of SSTEN embeddings can also be seen by comparing the top and bottom rows in Fig.~6 in the main manuscript. For example, by adding the temporal Gaussian kernel, we decrease the similarities in Fig.~6(a1) according to the temporal distance between two frames, which leads to clearer diagonal block structure in Fig.~6(a2).
Thus, we set $\sigma'=1/6$ for all following evaluations.

\subsection{Impact of Cluster Order}\label{sec:impact_cluster_order}
One merit of performing within-video clustering is that we can derive the temporal order of sub-clusters for each video separately. The video-wise individual order of clusters is used to guide the Viterbi decoding, which breaks the common assumption that clusters follow the same temporal order in all the videos. In the following, we verify the efficacy of the derived video-wise order of clusters. We use the same within-video clustering result with global cluster assignment and perform Viterbi decoding using two different temporal cluster orders: (1)~\textit{video-wise} order: the temporal order of sub-clusters is determined on each video separately; and (2)~\textit{uniform} order: the uniform order is determined by sorting the average timestamps of global clusters and is then applied to all the videos. Table~\ref{tab:impact_cluster_order_bf} reports the segmentation performance (after Viterbi) with these two orders for our SSTEN embeddings on Breakfast and YTI.
To measure the correctness of the predicted segment order, we adopt the segmental edit distance (Edit), which is a common metric for supervised action segmentation, \eg,~\cite{farha2019ms, lea2017temporal, morariu2011multi, koppula2013learning}. It penalizes segmentation results that have a different segment order than the ground truth (\ie,~it penalizes out-of-order predictions, as well as oversegmentation).

\begin{table}[tb]
\scriptsize
\centering
\caption{Impact of cluster order for two-step clustering on SSTEN embeddings (in \%).}
\begin{tabular}{M{0.7cm}M{0.3cm}M{0.3cm}M{0.2cm}M{0.3cm}M{0.3cm}M{0.3cm}M{0.2cm}M{0.3cm}}
\toprule
\multirow{2}*{Order} & \multicolumn{4}{c}{Breakfast} & \multicolumn{4}{c}{YTI} \\
\cmidrule(r){2-5}\cmidrule(r){6-9}
 ~ & MoF & IoU & F1 & Edit & MoF & IoU & F1 & Edit \\
\midrule
 video-wise &  50.3 &  \textbf{19.0} &  \textbf{33.6} &  \textbf{42.3} & 
\textbf{46.6} & \textbf{10.7} & \textbf{29.5} & \textbf{25.5}\\ 
  uniform &  \textbf{53.5} &  15.7 &  32.2 &  33.0 & 
40.7 & 7.7 & 25.1 & 20.3 \\

\bottomrule
\end{tabular}
\vspace{2mm}
\label{tab:impact_cluster_order_bf}
\end{table}

From Table~\ref{tab:impact_cluster_order_bf} we see that our video-wise order clearly outperforms the uniform order except for MoF on Breakfast.
Furthermore, the edit score verifies that our derived video-wise temporal orders are valid. 

In our experiments we especially notice that the MoF and IoU scores could act contradictory to each other, \eg,~the uniform order results in higher MoF scores (on Breakfast) at the cost of lower IoU scores. MoF tends to overfit on dominant classes (\eg, classes with longer action instances) while IoU is sensitive to underrepresented classes and penalizes segmentation results with dominating segments. Therefore, it is necessary that we consider all metrics for evaluation, as a higher MoF score does not always correspond to better performance in practice. 

\subsection{Impact of Decoding Strategies}\label{sec:impact_decoding_strategies}
We compare our approach, which uses Viterbi decoding, with the Mallow model decoding that has been proposed in \cite{sener18}. The authors propose a Rankloss embedding over all video frames from the same activity with respect to a pseudo ground truth action annotation. The embedded frames of the whole activity set are then clustered and the likelihood for each frame and for each cluster is computed. For the decoding, the authors build a histogram of features with respect to their clusters with a hard assignment and set the length of each action with respect to the overall amount of features per bin. After that, they apply a Mallow model to sample different orderings for each video with respect to the sampled distribution. The resulting model is a combination of Mallow model sampling and action length estimation based on the frame distribution.

For this experiment, we evaluate the impact of the different decoding strategies on two embeddings: the Rankloss embedding \cite{sener18} and our SSTEN embedding.
Table~\ref{tab:comparison_embeddig_decodings} reports the results of these two embeddings in combination with three decodings: the Mallow model, Viterbi decoding with K-means and Viterbi decoding with two-step clustering.

\begin{table}[!htb]
\scriptsize
\centering
\caption{Comparison of combinations of embeddings and decoding strategies on Breakfast (in \%).}
\begin{tabular}{ ccccccc}
\toprule
\multirow{2}*{Decoding}  & \multicolumn{3}{c}{Rankloss \cite{sener18} embed.} & \multicolumn{3}{c}{SSTEN embed.} \\
\cmidrule(r){2-4}\cmidrule{5-7}
~ &  MoF & IoU & F1 & MoF & IoU & F1 \\
\midrule
Mallow \cite{sener18} & 34.7 & \textbf{17.8} & \textbf{31.4} & 36.4 & 18.1 & 31.5 \\
kmeans+Viterbi & \textbf{35.2} & 15.6 & 28.8 &  39.3 & 17.8 & 31.9 \\
two-step.+Viterbi & 34.7 & 13.4 & 23.7 &   \textbf{50.3} & \textbf{19.0} & \textbf{33.6} \\
\bottomrule
\end{tabular}
\label{tab:comparison_embeddig_decodings}
\end{table}

Following \cite{sener18}, we run 7 iterations for the Rankloss embedding with the Mallow model. In each iteration, the Rankloss embedding is retrained using the segmentation result from the last iteration as pseudo label, and the frame-wise likelihoods and the Mallow model are updated. 

Unlike the Mallow model, our Viterbi decoding is a one-iteration procedure. It is operated on the embedding which is trained only once. When combining with Viterbi, we train the Rankloss model only once using the initialized uniform segmentation as a prior. For SSTEN with the Mallow model, we only run for one iteration, as we do not need to train SSTEN with pseudo labels iteratively. 

Considering the Rankloss results in Table~\ref{tab:comparison_embeddig_decodings} we see that combining it with the Mallow model achieves its highest IoU and F1 scores.
This is because for Viterbi decoding, the Rankloss model trained only one-time using the uniform initialization as pseudo label is lacking of a strong temporal prior.
Considering SSTEN, our Viterbi decoding with two-step clustering clearly outperforms the Mallow model. With Mallow, the SSTEN embedding has competitive IoU and F1 scores but significantly lower MoF. We also tried running the Mallow model on SSTEN embedded features for multiple iterations. However, this resulted in a reduced number of clusters. Thus, we see that an appropriate combination of embedding and decoding strategy is necessary.

\begin{figure}[!htb]
\begin{center}
    \includegraphics[width=0.85\linewidth]{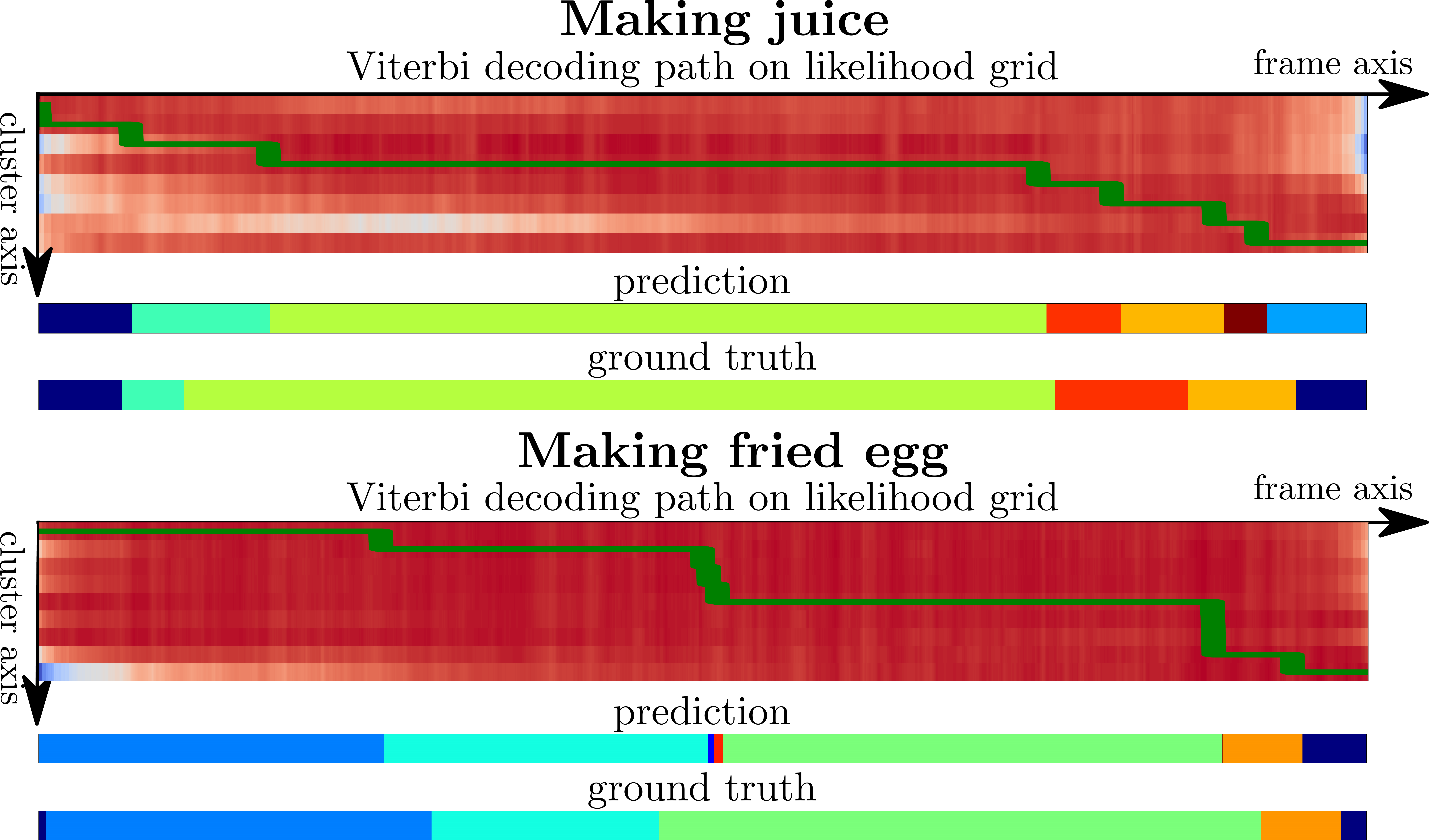}
\end{center}
  \caption{Comparison of Viterbi decoding path on the likelihood grid computed from the global clusters resulted from two-step clustering on the SSTEN embeddings, for two videos on Breakfast. The warm (red)/cool (blue) colors in the grid indicate high/low likelihoods of a frame belonging to an action class. It shows that the decoding assigns most frames to occurring actions while marginalizing actions that do not occur in the sequence by assigning only a few frames.
  }
\label{fig:exp_Viterbi_decoding}
\end{figure}

To have a closer look into the Viterbi decoding, we visualize the likelihood grids computed from global clusters,  
as well as the resulting decoding path over time for two videos on Breakfast in Fig.~\ref{fig:exp_Viterbi_decoding}. It shows that the decoding, which generates a full sequence of actions, is able to marginalize actions that do not occur in the video by just assigning only very few frames to those ones and the majority of frames are assigned to the clusters that occur in the video. 
Even if the given temporal order constrains that the resulting $K$ coherent segments have to follow the fixed temporal order, the segments that actually do not belong in the sequence will be marginalized because the Viterbi algorithm decodes a path that maximizes the posterior probability. 
Overall, it turns out that the Viterbi decoding constrained by a temporal order performs better than the Mallow model's iterative re-ordering. 

\subsection{Quantitative Segmentation Results}\label{sec:quantatitive_results}
\subsubsection{Results of Clustering and Final Segmentation}
In order to show the advantage of two-step clustering over K-means, when combined with the proposed SSTEN embedding, we report both, the results of clustering (before Viterbi decoding) and the final segmentation performance (after Viterbi decoding) on Breakfast in Table~\ref{tab:clustering_and_final}.
\begin{table}[t]
\scriptsize
\centering
\caption{Comparison of combinations of SSTEN and different clustering methods in terms of clustering and final segmentation after Viterbi decoding on Breakfast (in \%).}
\begin{tabular}{M{0.4cm}cM{0.2cm}M{0.2cm}M{0.2cm}M{0.2cm}M{0.2cm}M{0.2cm}}
\toprule
\multirow{2}*{Embedding} &  \multirow{2}*{Clustering} &  \multicolumn{3}{c}{Clustering results} &  \multicolumn{3}{c}{Final results} \\
\cmidrule(r){3-5}\cmidrule{6-8}
~ &  ~ &  MoF &  IoU &  F1 &  MoF &  IoU &  F1 \\
\midrule
\multirow{2}*{SSTEN} &  K-means & 27.2 & 13.5 & \textbf{26.3} & 39.3 & 17.8 &  31.9 \\ 
~ &  two-step.cluster & \textbf{38.6} & \textbf{13.7} & 25.9 & \textbf{50.3} &  \textbf{19.0} &  \textbf{33.6}  \\
\bottomrule
\end{tabular}
\vspace{2mm}
\label{tab:clustering_and_final}
\end{table}
We see that the proposed two-step clustering leads to superior performance than K-means, in terms of both clustering (before Viterbi decoding) in most metrics, and in terms of final segmentation (after Viterbi decoding).

\subsubsection{Segmentation Results On Each Activity}
We report the ground truth number of classes and segmentation performance of MLP with K-means (\emph{MLP+kmeans}, our reimplementation of \cite{kukleva19}) and TAEC for each activity on Breakfast (Table~\ref{tab:performance_bf}), YTI (Table~\ref{tab:performance_yti}) and 50 Salads (Table~\ref{tab:performance_fs}). The evaluation is done with global Hungarian matching on all videos. The number of clusters is set to the maximum number of ground truth classes for each activity ($K$ = max.\#gt). 

\begin{table}[!htb]
\scriptsize
\centering
\caption{Maximum number of ground truth action classes and segmentation performance of MLP+kmeans (our reimplementation of \cite{kukleva19}) and TAEC for the 10 activities on Breakfast (in \%).
The number of clusters is set to the maximum number of ground truth classes for each activity ($K$ = max.\#gt). }
\begin{tabular}{cM{0.8cm}cccc}
\toprule
\multicolumn{6}{c}{\textbf{Breakfast}} \\
\midrule
\multicolumn{6}{c}{Global Hungarian matching on all videos ($K$ = max.\#gt)} \\
\midrule
Activity  &  $K$ & Methods &  MoF & IoU & F1\\
\midrule
\multirow{2}*{coffee} & \multirow{2}*{7} & MLP+kmeans & \textbf{46.8} & \textbf{15.7} & \textbf{26.2}\\
~ & ~ & TAEC & 35.6 & 15.2 & 24.9 \\
\midrule
\multirow{2}*{cereals} & \multirow{2}*{5} & MLP+kmeans & 48.8 & 25.8 & 37.2\\
~ & ~ & TAEC & \textbf{59.0} & \textbf{31.4} & \textbf{47.7} \\
\midrule
\multirow{2}*{tea} & \multirow{2}*{7} & MLP+kmeans & 32.2 & 13.0 & 22.7\\
~ & ~ & TAEC & \textbf{39.2} & \textbf{16.3} & \textbf{26.1} \\
\midrule
\multirow{2}*{milk} & \multirow{2}*{5} & MLP+kmeans & 40.4 & 21.2 & 36.6\\
~ & ~ & TAEC & \textbf{46.7} & \textbf{27.3} & \textbf{43.5} \\
\midrule
\multirow{2}*{juice} & \multirow{2}*{8} & MLP+kmeans & 36.9 & 14.1 & 27.9\\
~ & ~ & TAEC & \textbf{52.2} & \textbf{22.3} & \textbf{36.2} \\
\midrule
\multirow{2}*{sandwich} & \multirow{2}*{9} & MLP+kmeans & 47.4 & 15.0 & 25.3\\
~ & ~ & TAEC & \textbf{53.7} & \textbf{19.8} & \textbf{33.5} \\
\midrule
\multirow{2}*{scrambled egg} & \multirow{2}*{12} & MLP+kmeans & 34.5 & 10.8 & 19.9\\
~ & ~ & TAEC & \textbf{48.1} & \textbf{15.2} & \textbf{28.3} \\
\midrule
\multirow{2}*{fried egg} & \multirow{2}*{9} & MLP+kmeans & 36.4 & 11.5 & 24.5\\
~ & ~ & TAEC & \textbf{49.1} & \textbf{17.4} & \textbf{30.5} \\
\midrule
\multirow{2}*{salad} & \multirow{2}*{8} & MLP+kmeans & 34.7 & 7.8 & 27.5\\
~ & ~ & TAEC & \textbf{42.0} & \textbf{15.2} & \textbf{34.0} \\
\midrule
\multirow{2}*{pancake} & \multirow{2}*{14} & MLP+kmeans & 57.4 & 8.6 & 19.2\\
~ & ~ & TAEC & \textbf{58.2} & \textbf{19.2} & \textbf{35.8} \\
\midrule
\midrule
\multirow{2}*{All} & \multirow{2}*{-}  & MLP+kmeans & 42.9 & 13.1 & 25.5\\
~ & ~ & TAEC & \textbf{50.3} & \textbf{19.0} & \textbf{33.6} \\
\bottomrule
\end{tabular}
\vspace{2mm}

\label{tab:performance_bf}
\end{table}

\begin{table}[!htb]
\scriptsize
\centering
\caption{Maximum number of ground truth action classes and the segmentation performance of MLP+kmeans (our reimplementation of \cite{kukleva19}) and TAEC for the 5 activities on YTI (in \%).
The number of clusters is set to the maximum number of ground truth classes for each activity ($K$ = max.\#gt). }
\begin{tabular}{M{0.5cm}M{0.3cm}cM{0.4cm}M{0.4cm}M{0.4cm}M{0.4cm}M{0.4cm}}
\toprule
\multicolumn{8}{c}{\textbf{YouTube Instructions}} \\
\midrule
\multicolumn{7}{c}{Global Hungarian matching on all videos ($K$ = max.\#gt)} \\
\midrule
Activity  & $K$ & Methods &  MoF w/o bkg & IoU w/o bkg & F1 w/o bkg & MoF w bkg & IoU w bkg\\
\midrule
\multirow{2}*{coffee} & \multirow{2}*{10} & MLP+kmeans & 40.9 & 10.4 & \textbf{34.2} & 11.9 & 9.5\\
~ & ~ & TAEC & \textbf{42.8} & \textbf{10.5} & 26.6 & \textbf{12.4} & \textbf{9.6} \\
\midrule
\multirow{2}*{change tire} & \multirow{2}*{11} & MLP+kmeans & 45.9 & 17.2 & 34.0 & 24.7 & 15.8\\
~ & ~ & TAEC & \textbf{58.6} & \textbf{20.0} & \textbf{37.2} & \textbf{31.5} & \textbf{18.4} \\
\midrule
\multirow{2}*{jump car} & \multirow{2}*{12} & MLP+kmeans & 30.6 & 4.5 & 24.4 & 5.1 & 4.1\\
~ & ~ & TAEC & \textbf{34.3} & \textbf{6.2} & \textbf{26.4} & \textbf{5.7} &  \textbf{5.8} \\
\midrule
\multirow{2}*{cpr} & \multirow{2}*{7} & MLP+kmeans & 34.4 & \textbf{9.9} & 31.2 & 15.0 & \textbf{8.6}\\
~ & ~ & TAEC & \textbf{38.0} & 7.9 & \textbf{33.3} & \textbf{16.6} & 6.9 \\
\midrule
\multirow{2}*{repot} & \multirow{2}*{8} & MLP+kmeans & 29.8 & \textbf{7.2} & \textbf{23.1} & 10.1 & \textbf{6.4}\\
~ & ~ & TAEC & \textbf{35.8} & 7.1 & 22.4 & \textbf{12.1} & 6.3 \\
\midrule
\midrule
\multirow{2}*{All} & \multirow{2}*{-}  & MLP+kmeans & 39.4 & 9.9 & \textbf{29.6} & 14.4 & 9.7 \\
~ & ~ & TAEC & \textbf{46.6} & \textbf{10.7} & 29.5 & \textbf{17.0} & \textbf{10.5}\\
\bottomrule
\end{tabular}
\vspace{2mm}
\label{tab:performance_yti}
\end{table}

\begin{table}[!htb]
\scriptsize
\centering
\caption{Maximum number of ground truth action classes and the segmentation performance of MLP+kmeans (our reimplementation of \cite{kukleva19}) and TAEC for the single activity on 50 Salads (in \%).
The number of clusters is set to the maximum number of ground truth classes for each activity ($K$ = max.\#gt). }
\begin{tabular}{cccccc}
\toprule
\multicolumn{6}{c}{\textbf{50 Salads}} \\
\midrule
\multicolumn{6}{c}{Global Hungarian matching on all videos ($K$ = max.\#gt)} \\
\midrule
Activity  & $K$ & Methods &  MoF & IoU & F1\\
\midrule
\multirow{4}*{salad} & \multirow{2}*{eval 12} & MLP+kmeans & 37.9 & 24.6 & 40.2 \\
~ & ~ & TAEC & \textbf{48.4} & \textbf{26.0} & \textbf{44.8} \\
\cmidrule{2-6}
~ & \multirow{2}*{mid 19} & MLP+kmeans & \textbf{29.1} & \textbf{15.7} & \textbf{23.4}\\
~ & ~ & TAEC & 26.6 & 14.9 & \textbf{23.4} \\

\bottomrule
\end{tabular}
\vspace{2mm}
\label{tab:performance_fs}
\end{table}

\subsection{Qualitative Segmentation Results}\label{sec:qualitative_results}
\begin{figure*}[!htb]
\centering
\includegraphics[width=\linewidth]{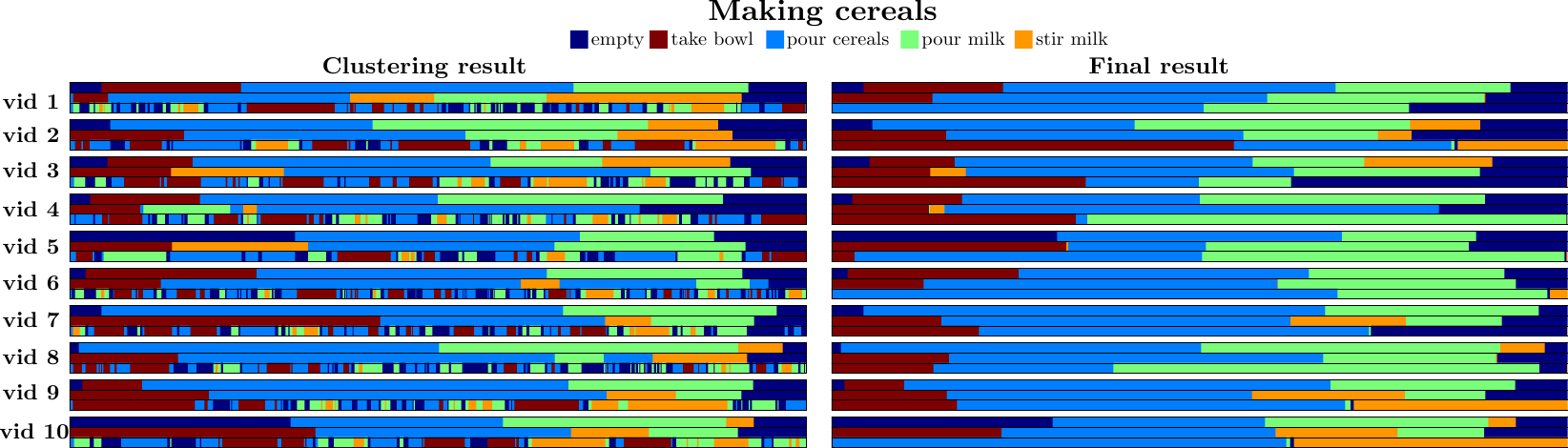}
\caption{Qualitative results of clustering and final segmentation of 10 \textit{making cereals} videos on Breakfast. For each video, the 3-row-group displays the ground truth (1st row), TAEC (2nd row), MLP+kmeans~\cite{kukleva19} (3rd row). The mapping between cluster labels and ground truth classes is done with global Hungarian matching on all videos. The number of clusters is set to the maximum number of ground truth classes for each activity ($K$ = max.\#gt).} 
\label{fig:segmentation_cereals}
\end{figure*}

\begin{figure*}[!htb]
\centering
\includegraphics[width=\linewidth]{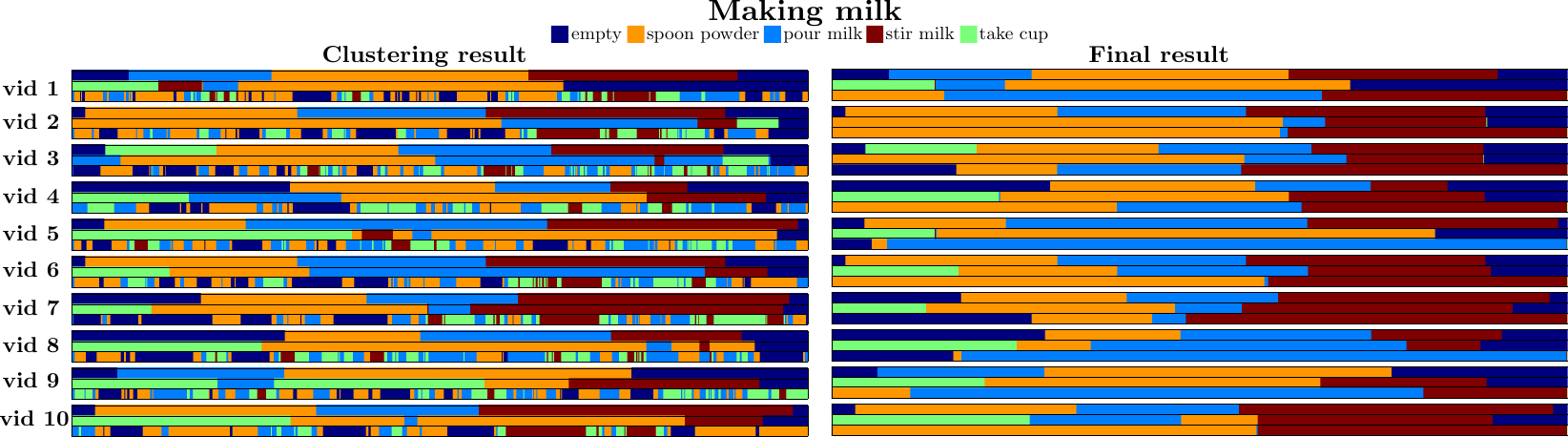}
\caption{Qualitative results for 10 \textit{making milk} videos on Breakfast.
} 
\label{fig:segmentation_milk}
\end{figure*}

\begin{figure*}[!htb]
\centering
\includegraphics[width=\linewidth]{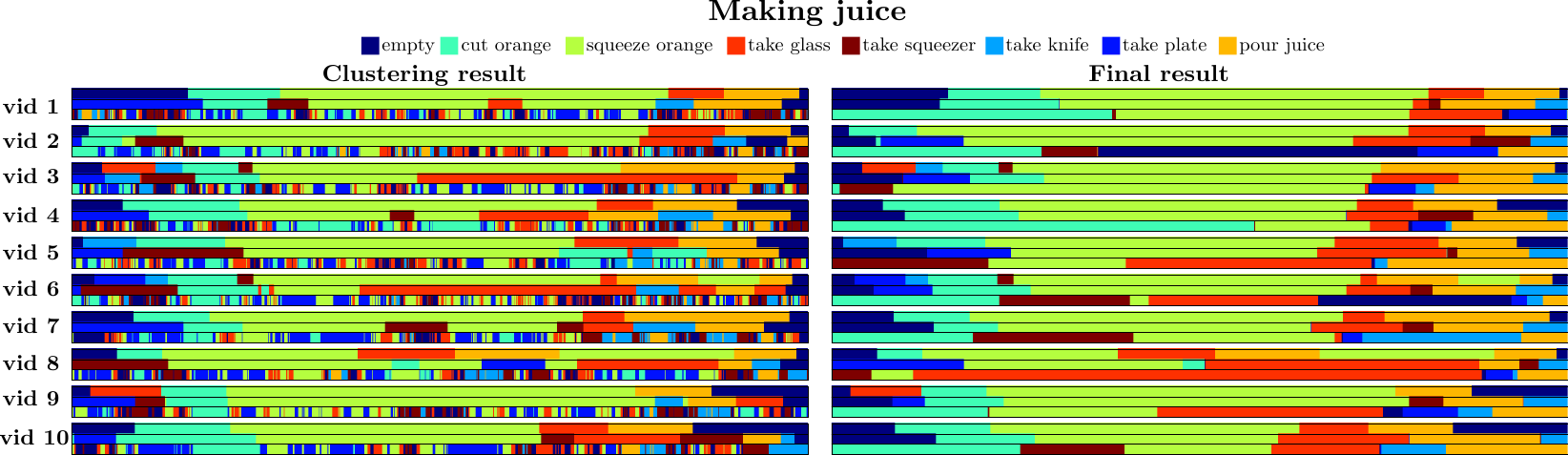}
\caption{Qualitative results for 10 \textit{making juice} videos on Breakfast.
} 
\label{fig:segmentation_juice}
\end{figure*}

\begin{figure*}[!htb]
\centering
\includegraphics[width=\linewidth]{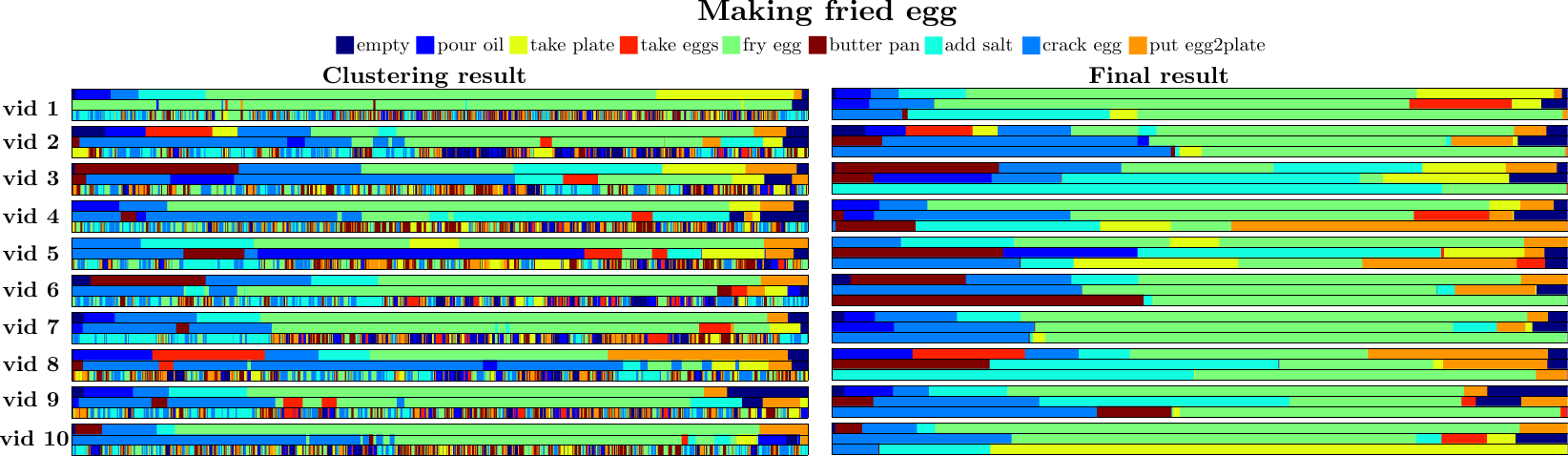}
\caption{Qualitative results for 10 \textit{making fried egg} videos on Breakfast.
} 
\label{fig:segmentation_friedegg}
\end{figure*}

\begin{figure*}[!htb]
\centering
\includegraphics[width=\linewidth]{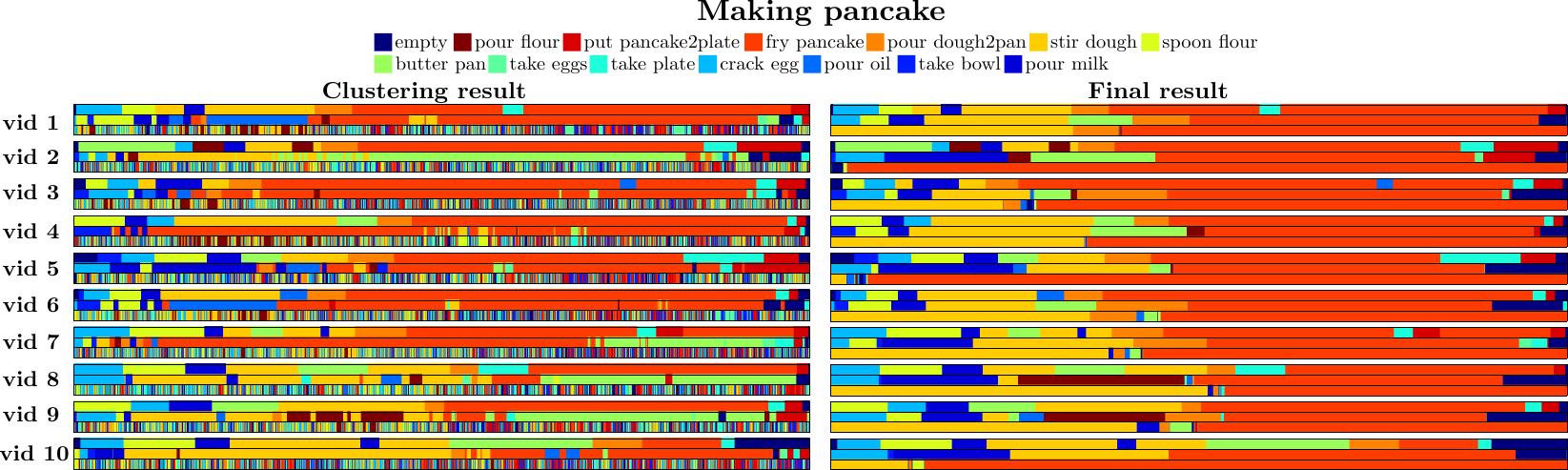}
\caption{Qualitative results for 10 \textit{making pancake} videos on Breakfast.
} 
\label{fig:segmentation_pancake}
\end{figure*}

\begin{figure*}[!htb]
\centering
\includegraphics[width=\linewidth]{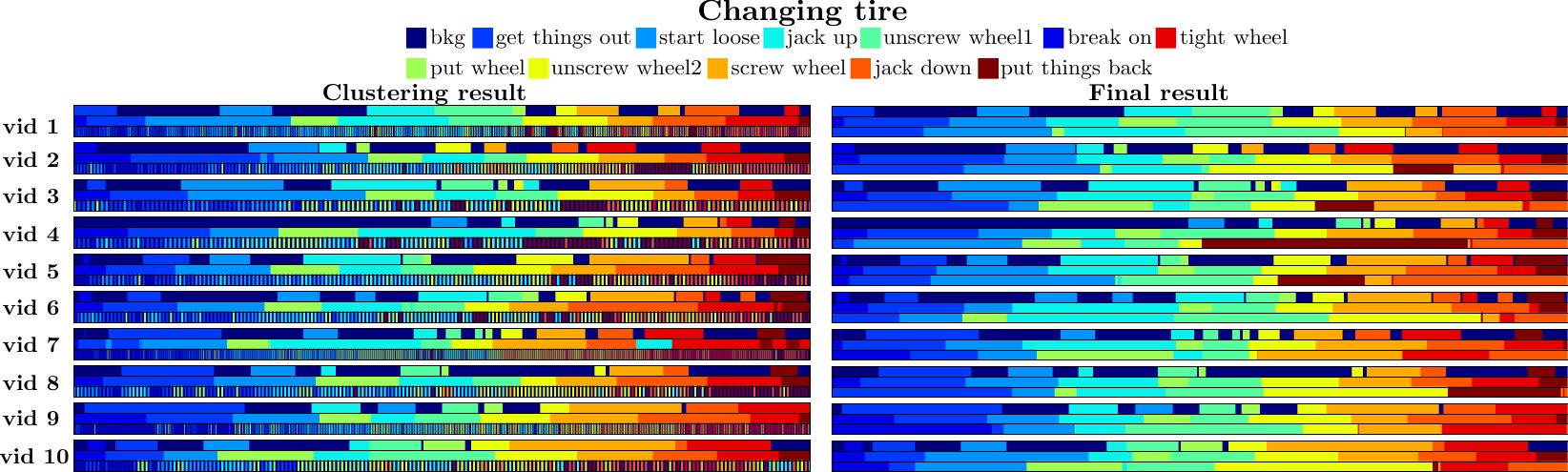}
\caption{
Qualitative results for 10 \textit{changing tire} videos of the YouTube Instructions dataset. Similar to the Breakfast illustrations, for each video the 3-row-group shows the ground truth (1st row), TAEC (2nd row), and MLP+kmeans~\cite{kukleva19} (3rd row).
} 
\label{fig:segmentation_changetire}
\end{figure*}

\begin{figure*}[!htb]
\centering
\includegraphics[width=\linewidth]{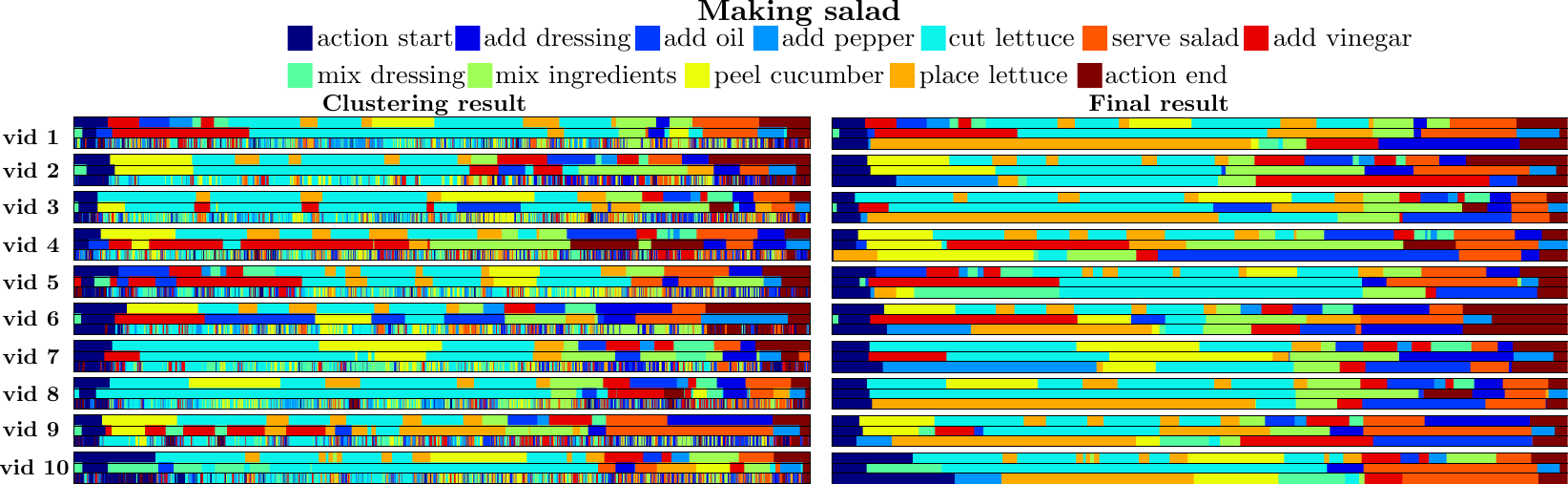}
\caption{Qualitative results for 10 \textit{making salad} videos on the 50 Salads dataset (from the \emph{eval}-level, 12 action classes).
Again, for each video the 3-row-group shows the ground truth (1st row), TAEC (2nd row), and MLP+kmeans~\cite{kukleva19} (3rd row).
} 
\label{fig:segmentation_fs}
\end{figure*}

We show the qualitative results of clustering and final segmentation on 7 composite activities: \textit{making cereals} (Fig.~\ref{fig:segmentation_cereals}), \textit{making milk} (Fig.~\ref{fig:segmentation_milk}), \textit{making juice} (Fig.~\ref{fig:segmentation_juice}), \textit{making fried egg} (Fig.~\ref{fig:segmentation_friedegg}), \textit{making pancake} (Fig.~\ref{fig:segmentation_pancake}) on Breakfast, \textit{changing tire} (Fig.~\ref{fig:segmentation_changetire}) on YTI and \textit{making salad} (Fig.~\ref{fig:segmentation_fs}) on 50 Salads (eval 12 classes). The mapping between cluster labels and ground truth classes is done with global Hungarian matching on all videos. The number of clusters is set to the maximum number of ground truth classes for each activity ($K$ = max.\#gt).

For each activity, we visualize the results of 10 videos. For each video, the 3-row-group displays the ground truth (1st row), TAEC (2nd row), MLP+kmeans~\cite{kukleva19} (3rd row). 

\bibliography{cvww23-bib}

\end{document}